\RequirePackage{fix-cm}
\documentclass[twocolumn]{svjour3}                     
\smartqed  
\usepackage{graphicx}
\usepackage[table]{xcolor}
\usepackage{tabularx}
\usepackage{array}
\usepackage{caption}
\usepackage{subfig}
\usepackage{amsmath}
\usepackage{amsfonts,amssymb}
\usepackage{color}
\usepackage{mathtools, cuted}
\usepackage{slashbox}

\advance\paperheight by-20mm
\usepackage[cam,cross,horigin=-27.9mm,vorigin=-13.5mm]{crop}

\usepackage{marvosym}
\newcommand{\envelope}{(\raisebox{-.5pt}{\scalebox{1.45}{\Letter}}\kern-1.7pt)}
\usepackage{times}
\usepackage[LY1]{fontenc}
\usepackage{biograph}

\smartqed

\begin{document}

\title{Sensorimotor Input as a Language Generalisation Tool
}
\subtitle{A Neurorobotics Model for Generation and Generalisation of Noun-Verb Combinations with Sensorimotor Inputs}

\titlerunning{Sensorimotor Input as a  Generalisation Tool}        

\author {Junpei Zhong \and {Martin Peniak} 
\and \\ {Jun Tani} \and  {Tetsuya Ogata} \and \\ Angelo Cangelosi }


\institute{J. Zhong \at Department of Intermedia Art and Science,  Waseda University, 3-4-1 Ohkubo, Shinjuku, Tokyo, Japan, 169-8555 \\ Centre for Robotics and Neural Systems, University of Plymouth, PL4 8AA,  United Kingdom \\
Tel.: +44 (0)175284908, 
\email{zhong@junpei.eu}\\
M. Peniak \at Cortexica Vision Systems, London, United Kingdom \\
J. Tani \at  Korea Advanced Institute of Science and Technology, Daejeon, South Korea \\
T. Ogata \at Department of Intermedia Art and Science,  Waseda University, Tokyo, Japan \\
A. Cangelosi \at Centre for Robotics and Neural Systems, University of Plymouth, United Kingdom
}

\date{Received: date / Accepted: date}

\maketitle

\begin{abstract}
The paper presents a neurorobotics cognitive model to explain the understanding and generalisation of nouns and verbs combinations when a vocal command consisting of a verb-noun sentence is provided to a humanoid robot. 
This generalisation process is done via the grounding process: different objects are being interacted, and associated, with different motor behaviours, following a learning approach inspired by developmental language acquisition in infants.
This cognitive model is based on Multiple Time-scale Recurrent Neural Networks (MTRNN).
With the data obtained from object manipulation tasks with a humanoid robot platform, the robotic agent implemented with this model can ground the primitive embodied structure of verbs through training with verb-noun combination samples. 
Moreover, we show that a functional hierarchical architecture, based on MTRNN, is able to generalise and produce novel combinations of noun-verb sentences. 
Further analyses of the learned network dynamics and representations also demonstrate how the generalisation is possible via the exploitation of this functional hierarchical recurrent network. 

\end{abstract}


\section{Introduction}

For the design of social robots~\cite{breazeal2004designing,dautenhahn2007socially}, besides of building robots with human-like external morphology, the ability to process, to understand and generate language is one of the key factors to support human-robot interaction.
However, to build a  model to accomplish similar processes for social robotics, the design of the robot's abilities of understanding, generation and generalisation of natural language is still  an open challenge. 
Particularly, natural language understanding for a social robotic system plays an essential role as it interfaces the vocal command from human users to an internal representation in the robot's own cognitive system. 
In this study we will follow a developmental robotics approach to the design of language and communication abilities in robots, following an incremental and interactive process to language learning, inspired by language development in infants.

\subsection{Language Understanding for Robot Systems}

Important recent developments in social robotics, such as robots performing human-like emotion expression~\cite{zhong2014continuous} and social attention for autonomous movement~\cite{novianto2014flexible}, 
have been accompanied by language understanding approaches focusing on the grounding of natural language into the agent's sensorimotor experience and its situated interaction~\cite{cangelosi2010grounding,steels2012language}.

For instance,
in~\cite{tellex2011understanding,matuszek2013learning}, syntactic parsing techniques are used to ground the language into primitive motor actions (e.g., pickup, move, place),  which can be inferred within graph models. 
Similarly, Misra et al.~\cite{misra2014tell} developed a system for mobile robots which is able to learn to ground the
language instructions from a corpus of pairs of natural language including both verbs and spatial information. 
~\cite{yuruten2013learning} proposed that in order to understand the object affordance which can be described by adjectives, the most crucial property is the shape-related one.

Besides  the direct modelling methods for robot language learning, an alternative approach to build a learning model for language is based on developmental robotics \cite{weng2001autonomous,asada2009cognitive,cangelosi2015developmental}.
Taking inspiration from developmental psychology and developmental neuroscience studies, this approach emphasises the role of the environment and of the interactions that occur  during learning, over a progression of learning stages.
In the context of language understanding, the core of developmental robotics approaches to language learning is following a similar developmental pathway of infants acquiring grounded representations of natural language and forming a symbol system through embodied interaction with the physical environment ~\cite{cangelosi2010integration}. 
Furthermore, via language learning an agent should also be able to generalise by inferring un-trained combinations of words within the lexical constructions acquired.

Various developmental robotics models have been developed that incrementally model the various stages of language acquisition in infants, from phoneme acquisition, to object and action names, to word combinations. For example, the cognitive model presented in~\cite{guenther2006cortical} outlines the cortical
interactions in the syllable generation process which result in different developmental phenomena. This mimics the first stage of language development.
The Elija model~\cite{howard2011modeling}  is a vocal apparatus which strictly follows  detailed developmental stages. Working as an articulatory synthesizer,  it firstly learns the production of sounds on its own. Then a caregiver is used to produce speech  by using speech sounds for object names using reinforcement learning, where the reward is again given by the response of the caregiver. 
Likewise, a self-organizing map together with reinforcement learning was proposed in~\cite{warlaumont2013prespeech}, which demonstrated  that the reinforcement learning based on the similarity of vocalization can improve the post-learning production of the sound of one's language.

From the  models mentioned above, we can see that most of the methods for modelling the first stages of phonetics production do not tend to use robotic platforms. 
On the other hand, for the modelling of the later stages of lexical development, 
after assuming that phonetics skills are mastered,  robotic systems are usually employed to establish the meta-knowledge about the association between vocal speech and the referents or the actions. Therefore, except studies focusing on the mental imagination of actions as in ~\cite{golosio2015cognitive}, the mechanical morphology of a robot is particularly important when modelling the acquisition of words, especially those used to name the motor actions. 
For instance, the model from~\cite{mangin2012learning} gets as input dance-like combinations of human
movement primitives plus ambiguous labels associated with these movements. 
Concentrating on the second and third stages of associating lexicon, words and motor actions, the robot in~\cite{dominey2009real} is able to acquire new motor behaviours
in an on-line fashion 
by grounding the vocal commands on the pre-defined control motor primitives. 
Similarly, Siskind~\cite{siskind2001grounding}  proposed a model which uses visual
primitives to encode notions of different actions to ground the
semantics of events for verb learning.
 Using structured connectionist models (SCMs), \cite{chang2005structured} built a layered connectionist model to connect embodied representations and
simulative inference for verbs.
In~\cite{cangelosi2004processing}, the emergence of verb-noun separation is learned while the agents are interacting and manipulating the objects.

~\cite{stramandinoli2012grounding} further developed the grounding the verbs with more complex meanings (such as ``keep'', ``reject'', ``accept'' and ``give'') which related to the internal states of the caregivers and which were used to build a robotic model for the grounding of increasingly abstract motor concepts and words.
As follow-up studies of~\cite{dominey2009real},~\cite{dominey2013recurrent,hinaut2013real} focused on the understanding of grammatical complexity. They used recurrent neural networks (RNN) to learn grammatical structure based on temporal series learning in artificial neural networks.
Also using RNN, \cite{sugita2005learning} reported experiments with a mobile robot implementing a two-level RNN architecture called Recurrent Neural Network with Parametric Bias Units (RNNPB). This allows the robot to map a linguistic command containing verbs and nouns into
context-dependent behaviours corresponding to  the verb and noun descriptions respectively. 
Comparing to RNNPB, another kind of RNN architecture called Multiple Timescale Neural Network (MTRNN) is able to ground different scales of sensorimotor information into the hierarchical structure of sentences, such as the spelling of words~\cite{ogata2013integration} and words and sentences~\cite{hinoshita2011emergence}. 
The kind of recurrent models provide a memory to store the spatial and temporal structure of the environment and the lexical structures. Given the fact that RNN can learn arbitrary length of the dependencies in  statistical structures and their context, the storage ability of the RNN   out-performs most of the language learning models.

On a higher level, concerning the meta-learning principles in these learning models, some developmental studies of language have focused on the intrinsic motivation to learn to speak, not only through reinforcement learning, but also following child psychology evidence that language learning can be driven without the explicit rewards from the caregivers. 
For instance, language commands can be acquired from learning from demonstration (LfD)~\cite{rohlfing2006can}, intrinsic plasticity~\cite{oudeyer2006discovering} and evolution~\cite{steels2005emergence,dautenhahn2012progress}.
These intrinsic motivation capabilities are implemented through learning models, which allow the agents to acquire
communication skills through vocal interactions,
besides the use of reinforcement learning techniques, 
such as the learning by demonstration model for grounding vocal commands into situated spatial information~\cite{forbes2015robot}; a comparative study for evolving robot language with situated information can be found in~\cite{parisi2002unified}.  
These models demonstrate different forces underlying language learning. 
From a mathematical prospective, 
those learning methods overcome the natural language learning bottlenecks of building compositionality of lexical structures and maximising the observed content~\cite{smith2003iterated} by means of statistical methods (LfD), optimal control (Reinforcement Learning) or other methodologies.

\subsection{Cognitive Background and Motivation}
In the developmental psychology studies focusing specifically on the learning of nouns and verbs, there  is still an open debate between the learning stages and their relative temporal acquisition order. 
For the early stages of verb and noun learning, it is widely accepted that  most of the common nouns are generally learned before  verbs~\cite{gentner1982nouns}, by first connecting speech sounds (labels, nouns) to physical objects in view.
However, some nouns which relate to context, such as ``passenger'', are learnt relatively late, only after ``an extensive range of situations'' (contexts or life phases) have been encountered~\cite{hall1993assumptions}, during which verbs may play a crucial role.
The embodied learning of verbs and nouns is not correlated to one single modality in sensory percept's: experiments done in \cite{kersten1998examination} suggest that the nouns are grounded from the intrinsic properties of an object, even at different movements and orientations, while verbs are  accounted for the movement path of an object. 
This distinction may be associated with the neuroanatomy distinction between the ventral and dorsal (what/where) visual streams, involved in the generation of nouns and verbs respectively. 
As~\cite{maguire200614} suggested, some nouns and verbs  can be learnt more straightforward to learn because they can be accessed perceptually. On the other hand, some  abstract words, either verbs or nouns, should only be learnt from a social and linguistic context. 
 
For instance, while  infants learn the word-gesture combination at the age of two, they associate the meaning of verbs with the meanings of the higher-order nouns~\cite{bates2002language}. Such verbs with complex meaning are obtained from both motor action and visual percept~\cite{longobardi2015noun}.
As summarised in~\cite{cangelosi2001nouns,Cangelosi2004}, comparing to the static object perception that associates to simple nouns, the early verb learning involves a temporal dynamic from motion perception.
Indeed, we assert that the learning processes of nouns and verbs (especially for those with complex meanings) are not separated; there is a close relation between verb and noun development, during which the embodied sensorimotor information plays a crucial role.
During this embodied development, 
both the perceptual system and the motor system contribute to language comprehension (e.g.~\cite{pulvermuller2002neuroscience,kaschak2005perception,pecher2003verifying,saygin2010modulation}).
These experiments also extend Piaget's proposal that language learning is a symbolised understanding process for dynamic actions, which is ``a situated process, function of the content, the context, the activity and the goal of the learner''.

The sensorimotor information is not the only mechanism acting as a learning tool for language acquisition. Conversely, recent research also proposes that language is such a flexible and efficient system for symbolic manipulation which is more than a communication tool of our thoughts (e.g.~\cite{landy2014perceptual,mirolli2009language,mirolli2011towards}.) 
For the predictive effect from language to sensorimotor behaviours, vocal communication can be one of the sources that drives the visual attention to become predictive, by making inferences as to the source-inferences~\cite{tomasello1986object}. 
In this process, language can trigger a predictive inference about the appearance of a visual percept, driving a predictive saccade~\cite{eberhard1995eye}. 
Therefore, the sensorimotor system is affected by the inferences from the auditory modality or even from higher level cognitive processes.

Following the hierarchical cognitive architecture proposed in~\cite{zhong2015artificial}, the language understanding can be
represented hierarchically from the neural processes on the (lower) receptor level to the higher level understanding which happens in the (higher) prefrontal cortex. 
Moreover, we will use a hierarchical recurrent neural architecture, as in~\cite{zhong2011robot,zhong2012learning,zhong2012predictive}, due to the fact that the learning modalities of visual perception and motor actions can be represented as both spatial and temporal sequences, so that the recurrent connections provide possibilities to intertwine these two modalities. 
In this paper, the MTRNN model will be employed to ground the features from different modalities with language structures in different time-scales.
Similar RNNPB~\cite{sugita2005learning} or MTRNN~\cite{heinrich2015analysing} networks have been used to learn verbs and nouns features with motor actions and visual features.  
The proposed model will use a single MTRNN model to learn both the sensory and motor information in a single set of sequences. 
We regard the perception and action having inseparable links (e.g.~\cite{wolpert1995internal,noe2001experience}) and should be encoded solely as similar data structures. 
Moreover, the training of such  a large MTRNN has become more and more feasible in recent years due to the accessibility and affordability of GPU computing . 
Therefore, the two modalities of our MTRNN can be conceptualised at the same time over the embodied sensorimotor experiences towards abstract and compact representations on the higher level of this hierarchy, similarly to the developmental processes of language conceptualisation and categorisation.

\section{The Multiple Timescale Recurrent Neural Network Model}

\begin{figure}
  \begin{center}
  \caption{Architecture of Multiple Timescale Recurrent Neural Network}
    \includegraphics[width=0.7\linewidth]{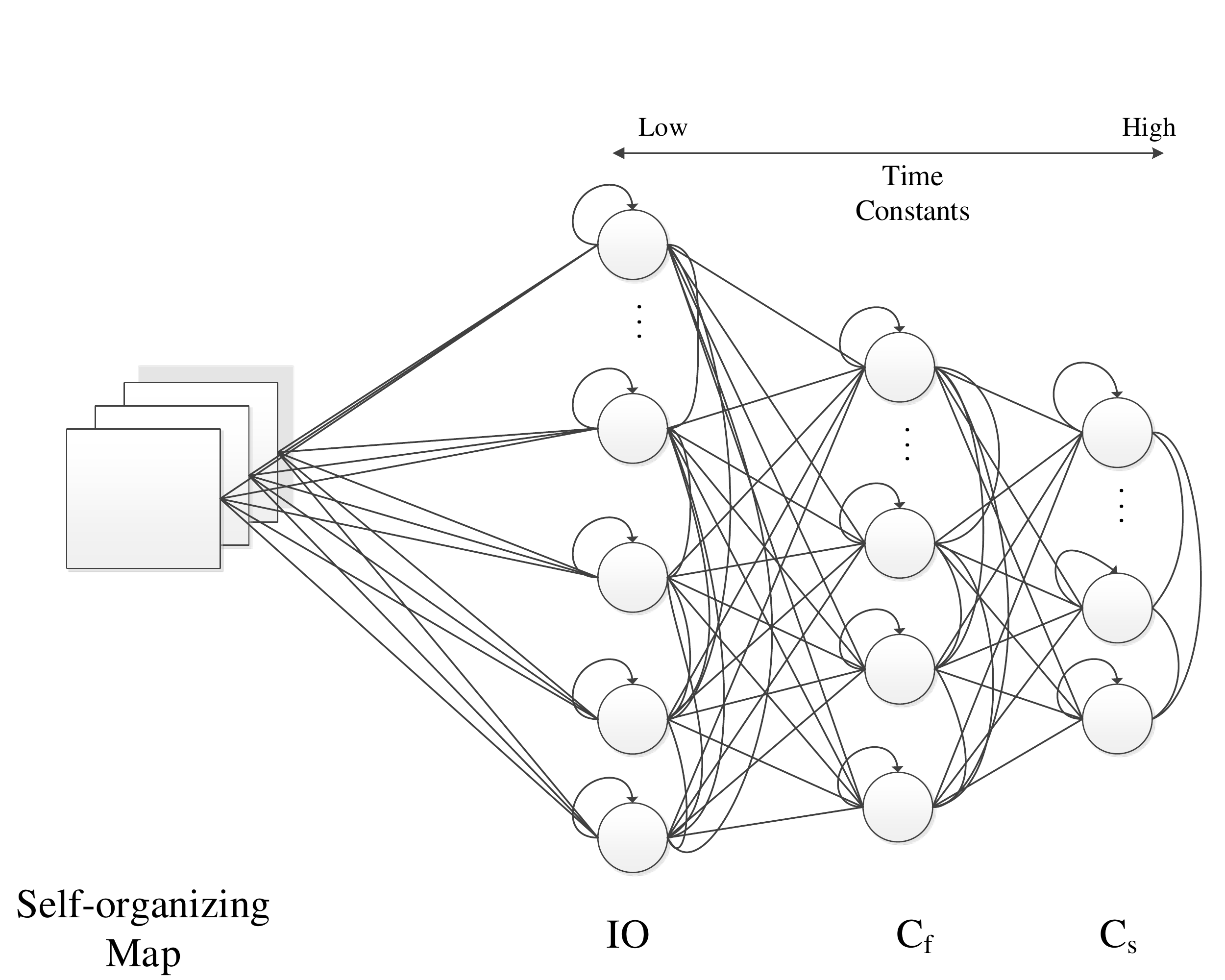}
    \label{fig:mtrnn_arc}
  \end{center}
\end{figure}

This model is based on the combination of a MTRNN network with Self-Organizing Maps (SOMs) to control the humanoid robot iCub, being trained on the understanding of a set of noun-verb combinations to perform a variety of actions with different objects. 
Fig.~\ref{fig:mtrnn_arc} shows the learning architecture incorporating a Multiple Timescale Recurrent Neural Network (MTRNN)~\cite{yamashita2008emergence} and the self-organizing maps.
The core module of the system is the MTRNN, which will learn sequences of verb-noun instructions and will control the movement of the robot in response to such instructions. The inputs to the MTRNN correspond to the language command inputs, to the visual inputs as well as the proprioceptive inputs.
We regard these three modalities as a whole sensorimotor input because the MTRNN model is able to learn the relation between verbs and nouns and seen objects within the context of the non-linearity of the sensorimotor sequences in a hierarchical manner. 
This network will learn this non-linearity in the functional hierarchy in which the neural activities  are self-organised,  exploiting the spatiotemporal variations.

\subsection{Using a Self-organizing Map as a  Sparse Structure}
The initial input data sets, consisting of speech, camera images, and kinaesthetic imitation proprioceptive states are pre-processed (See Eqs.~\ref{eq:som_input}-~\ref{eq:som_update}) using three SOMs respectively for the linguistic, visual and motor input modalities.  

Although the MTRNN could  be trained with original data representation, we usually employ pre-processing modules for the MTRNN inputs, which results in a sparse structure of the weighting matrices in the network. 
Also the MTRNN outputs are decoded into the original data structures. 
The sparseness in weighting matrices has a similar concept of sparse coding in computational
neuroscience~\cite{olshausen1997sparse}. 
The weighting matrices are sparsely distributed, which is an analogous form of the sparse distributed representations that are used in our brain, such as in visual~\cite{van1985functional} and auditory cortex~\cite{reale1980tonotopic}. 
Previous research on language learning in RNN~\cite{awano2011use} also showed that a sparse encoding results in robustness in training and a better generalisation results and improved robustness with noisy inputs.

Here the sparseness structure in the weight matrices is given by the SOMs~\cite{kohonen1998self}. 
During this process, the SOM performs as a dimensional mapping function, with an output space with higher dimensions than the input space.
Having a discretised and distributed neural encoding in the output space, the pre-processed SOM modules are able to reduce the possible overlap of the original data within the original input space. 
Therefore, the topological homomorphism produced by the SOM guarantees that the training vectors between the raw training-sets and the input vectors are topologically similar with each other.

In the SOM training here, assuming the input vectors are 

\begin{equation}
  x = [x^1 , x ^2 , ..., x ^m ]^\intercal 
  \label{eq:som_input}
\end{equation}

These input vectors  are mapped to an output space whose coordinates define the
output topology of the SOM.  Connecting between the input and output spaces, the weight vector is defined as

\begin{equation}
  w_j = [w_j^1 , w_j^2, \cdots, w_j^m ]^\intercal ,  j = 1, 2, 3, \cdots, n 
  \end{equation}

\noindent where  neuron $j$ is one of the input space vectors and 
$n$ is the total number of those neurons. 
When a self-organising map receives an
input vector, the algorithm finds a neuron associated with weights that are
most similar to the input vector. The measure of similarity is usually done
using the Euclidean distance metric, which is mathematically equivalent to
finding a neuron with the largest inner product $w^\intercal_j x$. 
Thus the very neuron that is the
most similar match for the input vector is referred to as best matching unit
(BMU) and it is defined as:

\begin{equation}
c = \mbox{arg } min_j  \Arrowvert x-w_j \Arrowvert  
\end{equation}

The dimensionality mapping is achieved when the  BMU
coordinates are used to update the weights of the neighbourhood neurons around neuron $c$ by driving them closer to the input vector at iteration $t$:
\begin{equation}
  w_j(t+1) = w_j(t) + \delta (x_j - w_j)
  \label{eq:som_update}
\end{equation} 
\noindent where $\delta$ is a neighbourhood function from the distance from BMU.

Therefore, the output of the SOM which is encoded in a high-dimensional
input space,   is still able to preserve the topological properties of the input space due to the use of the neighbourhood function.

\subsection{Multiple Timescale Recurrent Network (MTRNN)}

\begin{figure}[th]
	\begin{center}
		\caption{Language Learning Model based on MTRNN}
		\includegraphics[width=0.7\linewidth]{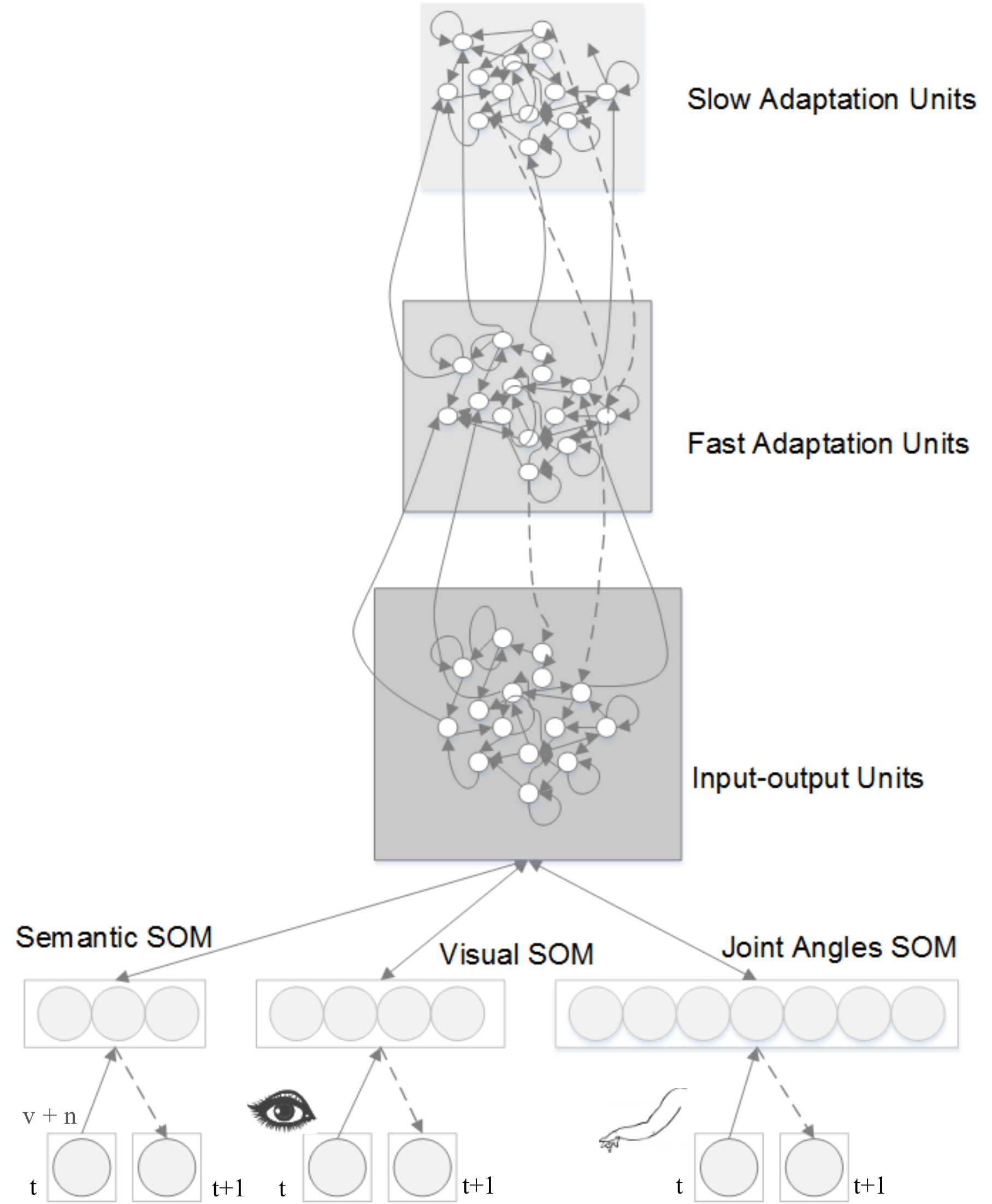}
		\label{fig:mtrnn_model}
	\end{center}
\end{figure}

As shown in Fig.~\ref{fig:mtrnn_model}, the neurons in the MTRNN form three layers: an input-output layer ($IO$) and two context layers called Context fast ($C_f$) and Context slow ($C_s$). 
In the following text, we denote the indices of these neurons as: 
\begin{equation}
  I_{all} = I_{IO} \cup I_{C_{f}} \cup I_{C_{s}}
\end{equation}

\noindent where $I_{IO}$ represents the indices to the neurons at the input-output layer, $I_{C_f}$ belongs to the neurons at the context fast layer and $I_{C_s}$ belongs to the neurons at the context slow layer.
The neurons on a layer own full connectivity  to all neurons within the same and  adjacent layers, as shown in Fig.~\ref{fig:mtrnn_arc}.
The difference between the  fast and slow context layers as well as the input-output layer consists in having distinct time constants $\tau$, which determine the speed of the adaptation given a time sequence with a specific length, when updating the neural activity. 
The larger the value of $\tau$, the slower the neuron adaptation.
The difference of adaptation rate of the neurons further assemble features of the input sequences in various timescales. 
Therefore, given the previous states $S(0), S(1), ..., S(t)$, their spatiotemporal features will be self-organised on different levels of the network.
So the MTRNN is not only a continuous time recurrent neural network that can predict the next states $S(t+1)$ of the time sequence, but also its internal state acts as a hierarchical memory to preserve the temporal features of the non-linear dynamics in different timescales. 

\subsubsection{Learning}

In general, the training of the MTRNN follows the updating rule of classical firing rate models, in which the activity of a neuron is determined by 
the average firing rate of all the connected neurons. 
Additionally, the neuronal activity is also decaying over time following an updating rule of leaky integrator model. 
Therefore, when time-step $t>0$,  the current membrane potential status of a neuron is determined both by the previous activation as well as the current synaptic inputs, as shown in Eq.~\ref{eq:neuron_status}:

\begin{equation}
  \tau_i u^{'}_{i,t} = -u_{i,t} + \sum_j w_{i,j} x_{j,t}
  \label{eq:neuron_status}
\end{equation}

\noindent where $u_{i,t}$ is the membrane potential, $x_{j,t}$ is the activity of $j$-th neuron at $t$-th time-step, $w_{i,j}$ represents the synaptic weight from the $j$-th neuron to the $i$-th neuron and $\tau$ is the time scale parameter which determines the decay rate of this neuron. 
One of the features that is similar to the generic continuous time recurrent neural networks (CTRNN) model is that a parameter $\tau$  is used to determine the decay rate of the neural activity; a larger $\tau$ means their activities change slowly over time compared  with those with a smaller $\tau$.

Assuming the $i$-th neuron has the number of $N$ connections (i.e. the total number of the neurons in the network is $N$), Eq.~\ref{eq:neuron_status} can be transformed into 

\begin{equation}
  u_{i, t+1} = (1 - \frac{1}{\tau_i}) u_{i,t} + \frac{1}{\tau_i}[\sum_{j \in N} w_{i,j} x_{j,t} ] \ \ (\mbox{if } t>0)
\end{equation}

When the time-step $t = 0$, the membrane potential of the $IO$ neurons is set to $0$ and the context neurons are set to initial states:

\begin{eqnarray}
     u_{i,0} = \left\{
\begin{aligned}
0, && \mbox{if }  t =0 \mbox{ and } i \in I_{IO}, \\
C_{{s_c}({i,0})}, &&  \mbox{if } t \neq 0 \mbox{ and } i \notin I_{IO} 
\end{aligned}
\right.
\label{eq:neuro_act1}
\end{eqnarray}

The neural activity of a neuron is calculated in two methods, depending on which level the neuron belongs with:

\begin{eqnarray}
     y_{i,t} = \left\{
\begin{aligned}
\frac{e^{u_{i,t}}}{\sum_{j \in Z} e^{u_{j,t}}}, && \mbox{if }  i \in I_{IO}, \\
\frac{1}{1+e^{-u_{i,t}}}, && \mbox{otherwise. }
\end{aligned}
\right.
\label{eq:neuro_act2}
\end{eqnarray}

\noindent Thus there is a sigmoid activation function for context neurons, while the input-output neurons are calculated by the soft-max function. 
The soft-max activation function gives rise to the recovery of a similar probability distribution as the SOM pre-processing modules.  Therefore, this activation function results in a faster convergence to the MTRNN network training.  

During training, the neurons of MTRNN self-adapt their weight matrices 
as well as the internal states of the neurons on the context layers for the processing of the incoming time sequence. 
The purpose of the training is to minimize the error $E$ which is defined by the Kullback-Leibler divergence in this case:\\

\begin{equation}
  \label{eq:energy}
   E = \sum_t \sum_{i \in O} y^{*}_{i,t} log(\frac{y^*_{i,t}}{y_{i,t}})  
\end{equation}

\noindent where $y^*_{i,t}$ is the desired neural activation of the $i$-th neuron at the $t$-th time-step, which acts as the target value for the actual output $y_{i,t}$. The target of the training is to minimize $E$ by back-propagation through time (BPTT).

In the BPTT algorithm, the input of the $IO$ neuron is calculated from a mixed partition value $r$ (called the feedback rate) of the previous output value $y$ and the desired value $y^*$. (Eq.~\ref{eq:feedback-rate})

\begin{equation}
x_{j,t+1} = (1-r) \times y_{j,t} + r \times y_{j,t} ^ *
\label{eq:feedback-rate}
\end{equation}

\noindent where we will use $r=0.1$ during training, and $r=0$ during generation, which means that the network is used to generate the sequences autonomously. 

At the $n$-th iteration of training, the synaptic weights and the biases of the network of neuron  $i$ are updated according to Eq.~\ref{eq:update_w}.

	
\begin{eqnarray}
\label{eq:update_w}
 w_{i,j}^{n+1} &=& w_{i,j}^n - \eta_{i,j} \frac{\partial E}{\partial w_{i,j}} \nonumber \\
&=& w_{i,j} -  \frac{\eta_{i,j}}{\tau_i} \sum_t  x_{j,t} \frac{\partial E}{\partial w_{i,t}}   \\
\label{eq:update_b}
 b_i^{n+1} &=& b_i^n - \beta_i \frac{\partial E}{\partial b_i} = b_i - \beta_i \sum_t \frac{\partial E}{\partial u_{i,t}}
\end{eqnarray} 

\begin{strip}
\begin{equation}
\label{eq:partial_diff}
  \frac{\partial E}{\partial u_{i,t}}= \left\{
\begin{aligned}
y_{i,t+1}-y^*_{i,t+1}+(1-\frac{1}{\tau_i})\frac{\partial E}{\partial u_{i,t+1}}, && \mbox{if }  i \in I_{IO}, \\
\sum_{k \in I_{all}} \frac{\partial E}{\partial u_{i,t+1}} [\lambda_{i,k}(1-\frac{1}{\tau_i}) + \frac{1}{\tau_k} w_{ki} \int'(u_{i,t}) ], && \mbox{otherwise. }
\end{aligned}
\right.
\end{equation}
\end{strip}

In Eq.~\ref{eq:update_w} and Eq.~\ref{eq:update_b},  the partial derivatives for $w$ and $b$ are the sums of weight and bias which determine the changes
over the whole sequence respectively, and $\eta$ and $\beta$  denote the learning rates for the weight and bias changes. 
Particularly, the term $ {\partial E}/{\partial u_{k,t}}$ can be calculated recursively as Eq.~\ref{eq:partial_diff}, where the $\int'()$ is the derivative  of the Sigmoid Function defined by Eqs.~\ref{eq:neuro_act1} and \ref{eq:neuro_act2}. The term $\lambda_{i,k}$ is the Kronecker's Delta, whose output is $1$ when $i = k$, otherwise it is set to $0$.

\section{Experiments}

To examine the network performance, we recorded the real world training data from object manipulation experiments based on an iCub  robot~\cite{metta2008icub}.
This is  a child sized  humanoid robot which was built as a testing platform for theories and models of cognitive science and neuroscience.  
Mimicking a two-year old infant, this unique robotic platform has 
 $53$ degrees of freedom.
As such, using the iCub, we set a learning scenario  in which a human instructor was teaching the
robotic learner a set of language commands whilst providing kinaesthetic demonstration of the named actions.   
The aim of these experiments was to evaluate the error for generalisation with a large data-set. We were also interested in the mechanisms, especially the neural activities in the hierarchical architecture, which result in such a generalisation. 

\subsection{Experimental Setup}

\begin{figure}
\begin{center}
\caption{Experimental Scenario}
  \includegraphics[width=0.7\linewidth]{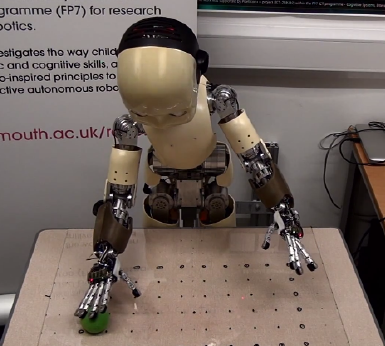}
	\label{fig:icub-setting}
	\end{center}
	\end{figure}

Fig.~\ref{fig:icub-setting} shows the setup used in our experiments. The data set was obtained using the following steps:

\begin{enumerate}
  \item Objects with significantly different colours and shapes were placed at $6$ different locations in front of the iCub. 
  \item A vocal command was spoken by an instructor according to the visual scene that was perceived by the iCub. A complete sentence of the vocal command is composed of a verb and a noun.
For instance, assuming we have the command ``lift [the] ball'', this was recognised by the speech recognition software called Dragon dictate\footnote{http://www.nuance.co.uk/dragon/index.htm}, with which the corresponding verb and noun were recognised and then translated into two dedicated discrete values based on the verb and noun dictionaries (Tab.~\ref{tab:verbnounlist}).
	\item The built-in vision tracker of the iCub searches for a ball-shaped object based on the dictionary-generated values; the iCub uses its
vision tracker system which incorporates visual segmentation algorithm to track a particular type of object, rotate the joints of head and neck and locate it in the visual field.
\item Once the object is located, the iCub rotates its head and triggers the object tracking,
which will change the encoder values of the neck and eyes.
	\item Joint positions of the head and neck are recorded. The sequence recorder module of the iCub was used to record the
sensorimotor trajectories while the instructor was guiding the robot by
holding its arms to perform a certain action for each object.
\item The hand and torso joints rotate  to certain angles to accomplish the lifting action toward the ball (with human instructor during training/without human instructor during execution)

\end{enumerate}

The whole experimental setup used   combinations of $9$ actions and $9$ objects.
From these combinations, both the vocal commands (i.e. a complete sentence includes verb and noun) and the sensorimotor sequences can be created. 
 To the best of our knowledge, this $9 \times 9$  noun-verb scenario is one of setups with the highest combination of verbs and nouns in grounded robot language experiments (e.g.~\cite{tani2004self,yamashita2008emergence})
We used such a large number of data  to test the combinatorial complexity and mechanical feasibility of this model, as well as to evaluate the generalisation ability and its internal non-linear dynamics when using such a large data-set. 
From an engineering point of view, after testing the feasibility of generalisation, it is also possible to apply this model in a real-world robot application. 

\begin{table}[ht]
\caption{Dictionaries of verbs and nouns for the data sets: The instructor showed the robot with different combinations from the $9$ actions and $9$ nouns. The  actions and the objects are represented in two discretised values for semantic command inputs which range from $0-0.9$. For instance, the command ``lift [the] ball'' is translated into values $[0.8, 0.2]$. }

\begin{center}
  
\begin{tabular}{|c||c|c|c|c|c|c|}
\hline
Actions & Slide Left & Slide Right & Touch & Reach & Push \\

\hline
Verb Value    & 0.0 & 0.1 & 0.2 & 0.3 & 0.4  \\
 
\hline
\hline
Actions &   Pull & Point & Grasp & Lift & \\

\hline
Verb Values   & 0.5 & 0.6 & 0.7 & 0.8 & \\

\hline
\hline 
\hline
Objects & Tractor & Hammer & Ball & Bus & Modi \\
\hline
Noun Value    & 0.0 & 0.1 & 0.2 & 0.3 & 0.4  \\
\hline
\hline
Objects &  Car & Cup & Cubes & Spiky & \\
\hline
Noun Values   & 0.5 & 0.6 & 0.7 & 0.8 & \\
\hline

	\end{tabular}
\label{tab:verbnounlist}
\end{center}
\end{table}

As mentioned before, each speech command was recognised and translated into two semantic command units.  Using $9$ discretised values for verbs and $9$ for nouns, the semantic commands have thus $81$  possible combinations. 
This translation was done according to the verb and noun dictionaries, as shown in  Tab.~\ref{tab:verbnounlist}. 
Since we used the visual object tracker in the iCub, the joints of neck and eyes automatically represent the location of  the particular object which is presented in the vocal commands. 
 Also the movements of the joint angles in the torso are recorded as the sequences of the motor actions. 
During the data recording, each recording sequence lasted $5$ second and the encoder values of $41$ joints were sampled at $50ms$ intervals.
Thus, the complete input vector of the data set  
contains $100$ temporal steps of the discrete semantic command, location of  visual attention and joint movement of the torso, as shown in Tab.~\ref{tab:data}.

\begin{table*}[ht]
		\begin{center}
		\caption{Structure of the Training Data}
	
		\begin{tabular}{|c||c|c|c|}
			\hline
			Description & Semantic Commands  & Object Location (Neck and Eyes) & Torso  Joints\\
			\hline
			Dimension & 2 & 6 & 3 \\
			\hline 
			Description &  Left Arm Joints & Right Arm Joints & \\
			 \hline
			 Dimension & 16 & 16 & \\
			 \hline
		\end{tabular}
		\label{tab:data}
	\end{center}
\end{table*}

Three experiments were carried out and are described in the next subsections: in the first experiment, given the 9 actions and 9 objects data set, we will search the parameter space and find the best parameters for the network training. 
In the second experiment, the training and generalisation performance will be shown given different types of manipulated data sets. 
For the third experiment, we will further analyse the generalisation ability of the MTRNN network. 
All these experiments were run using a modified version of the Aquila software ~\cite{peniak2011aquila} in a GPU computer with one Tesla C2050 and two GeForce GTX 580 graphic cards.
   
\subsection{Experiment 1 - Training Setting}

During  this section, we  used the data set consisting of the complete $9 \times 9$ combinations (i.e. $N_v = 9$, $N_n=9$), which include information about $6$ different object locations. Thus the whole data-set contains $9 \times 9 \times 6 = 486$ sequences, which were all used for training the network.
The most distinct feature  of the MTRNN, with respect to generic RNN or CTRNN networks, is that different neurons have distinct time constants $\tau$ , which are also one of the key factors that determine the training performances of the network. 
In this experiment, we  systematically changed these parameters in the parameter space in order to find out the best parameter settings.

\begin{table*}[th]
	\begin{center}
	\caption{Training Error with Different Parameter Settings $(C_s, C_f, N_{C_s}, N_{C_f})$}
	\label{tab:parameters-tuning}
	\begin{tabular}{|c|c|c|c|c|}
		\hline
			Parameters  & Error 1 & Error 2 & Error 3 & Ave. \\
			\hline 
			(70, 5, 20, 60) & 0.084	& 0.081 &	0.085 & 0.0833  \\
(70, 3, 20, 60)	& 0.084	& 0.085	& 0.082 &	0.0837\\
(70, 5, 30, 60)	& 0.084	& 0.086 &	0.083 & 0.0843	\\
(70, 5, 30, 50)		&	0.082	& 0.079	& 0.080 & 0.0803	\\
(70, 5, 30, 100)	& 0.079	& 0.079 &	0.078 & 0.0787	\\
(70, 5,  60, 100) &	0.078	 &0.078	 & 0.077 & 0.0777	\\
(70, 5,  40, 120) &	0.079	 &0.079	 & 0.078 & 0.0787	\\
(70, 5,  50, 140) &	0.075	 &0.075	 & 0.077 & 0.0757	\\
(70, 5, 60, 160)	& 0.072	& 0.071	& 0.074	& 0.0723\\
\textbf{(70, 5, 50, 120)}	& \textbf{ 0.071	}	& \textbf{ 0.070	}	& \textbf{ 0.071	}	& \textbf{ 0.0707}\\
\textbf{(70, 3, 50, 120) }	& \textbf{	0.071	}	& \textbf{ 0.071	}	& \textbf{ 0.070	}	& \textbf{ 0.0707}\\
 (70, 5, 70, 120)		&  { 0.070	}	&  { 0.071	}	& { 0.072	}	&  { 0.0710}\\
\hline
			\end{tabular}
	\end{center}
\end{table*}

A parameter space is defined as $( \tau_s, \tau_f, N_{C_s}, N_{C_f} )$, representing the time constants on the context slow layer, context fast layer and the number of neurons on these two layers respectively. 
In order to minimise the effects of the randomness of the initialisation, a total of $3$ training trials were done with the same training setting, as shown in  Tab.\ref{tab:parameters-tuning}.  
Previous MTRNN experiments in the literature have reported different parameters settings (\cite{yamashita2008emergence}, \cite{heinrich2012adaptive} and~\cite{hinoshita2009emergence}). 
From these experiments, we can discover that the number of neurons on the  $C_s$ and $C_f$ layers were determined mainly according to dimension of the $IO$ layer, but they generally kept a ratio  from $1:4$ to $1:3$.
To start, we firstly set the parameters according to the minimum values $(70, 5, 20, 60)$ from previous research~\cite{yamashita2008emergence}. 
Then we  scaled up the numbers of neurons on the context layers and adjusted the time constants. 
The less crucial parameters were kept constantly: learning rates $\eta = 0.7$, $\beta = 0.7$, momentum = $0.9$, weightRange = $0.025$. 
The stopping criteria for the training process was that the error did not decrease more than $1^{-6}$ within consecutive $100$ iterations. 
From Tab.~\ref{tab:parameters-tuning}, we can see that the number of neurons on the context layers affected much on the training performance of the network. 
Comparatively, the time constants played a less significant role for the training error than the number of neuron did. 
Also results showed that the suitable ratio for numbers of $C_s$ and $C_f$ neurons should be kept to around $1:4$ to $1:3$.


\begin{table*}[ht]
\begin{center}
	
		\caption{Some of the sequences containing particular semantic combinations of verbs and nouns were removed during training. The number $i$ in the cell indicates that such a combination was removed in the $i$-th training set for generalisation experiments. }
	\begin{tabular}{|c|c|c|c|c|c|c|c|c|c|}
		\hline
		\backslashbox{V.}{N.} & 0.0&0.1 & 0.2 & 0.3&0.4&0.5&0.6&0.7&0.8 \\ 
		\hline
		0.0& {\color{red}1}/{\color{blue}2}/{\color{green}3}  & &  & {\color{green}3} & & {\color{blue}2} & {\color{green}3} && \\
		\hline
		0.1&  {\color{blue}2}/{\color{green}3} & {\color{red}1} &  & {\color{green}3}& & {\color{blue}2}&{\color{green}3} && \\
		\hline
		0.2& {\color{green}3} & {\color{blue}2} & {\color{red}1} & {\color{green}3}& & &{\color{blue}2}/{\color{green}3} && \\
		\hline
		0.3&  & {\color{blue}2}/{\color{green}3}  &  &  {\color{red}1} & {\color{green}3}& & {\color{blue}2} &{\color{green}3}& \\
		\hline
		0.4&  &  {\color{green}3} & {\color{blue}2} & &  {\color{red}1}/{\color{green}3} & & &{\color{blue}2}/{\color{green}3}& \\
		\hline
		0.5&  &  {\color{green}3} & {\color{blue}2} & & {\color{green}3} & {\color{red}1} & & {\color{blue}2}/{\color{green}3}& \\
		\hline
		0.6&  &   & {\color{green}3} &{\color{blue}2} & & {\color{green}3}&  {\color{red}1} && {\color{blue}2}/{\color{green}3}\\
		\hline
		0.7&  &   & {\color{green}3} & {\color{blue}2}& & {\color{green}3}& & {\color{red}1} &{\color{blue}2}/{\color{green}3} \\
		\hline
		0.8&  &   & {\color{green}3} & &{\color{blue}2} & {\color{green}3}& &&{\color{red}1}/{\color{blue}2}/{\color{green}3}\\
		\hline
	\end{tabular}
	\label{tab:remove_list}
  \end{center}
\end{table*}

\subsection{Experiment 2 - Training Performance}
From the previous experiment,  we found that the best parameters  for training the $9 \times 9$ verb-noun data-set are those shown in bold in Tab.~\ref{tab:parameters-tuning}, among which we selected $(50,5,70,100)$.  
We then  examined the training performance of the network under this parameter setting using different data-sets.  
To test the generalisation ability, these data-sets were manipulated: a subset of the combinations of actions and objects were removed from the training set, to be used  as validation test sets when testing the generalisation ability of the network. 
The detailed information about the manipulated data-sets are shown in Tab.~\ref{tab:remove_list}, where the coloured numbers $N$ indicate the specific verb-noun combination  removed in the specific $N$-th data-set. 
We can see that the number of removal sets were increasing from the first to the third test-set, indicating the difficulty of generalisation was increasing. Also at the second and the third data-sets, some of the removal sets were next to each other, which further increased the difficulty of generalisation.

\begin{figure*}[ht]
  
    \caption{Training Curves with Three Test-sets}
    \label{fig:three-training-curves}
    \centering
    \subfloat[Training Curve with Test-set 1\label{fig:training_curve_test2}]
   {\includegraphics[width=0.5\textwidth]{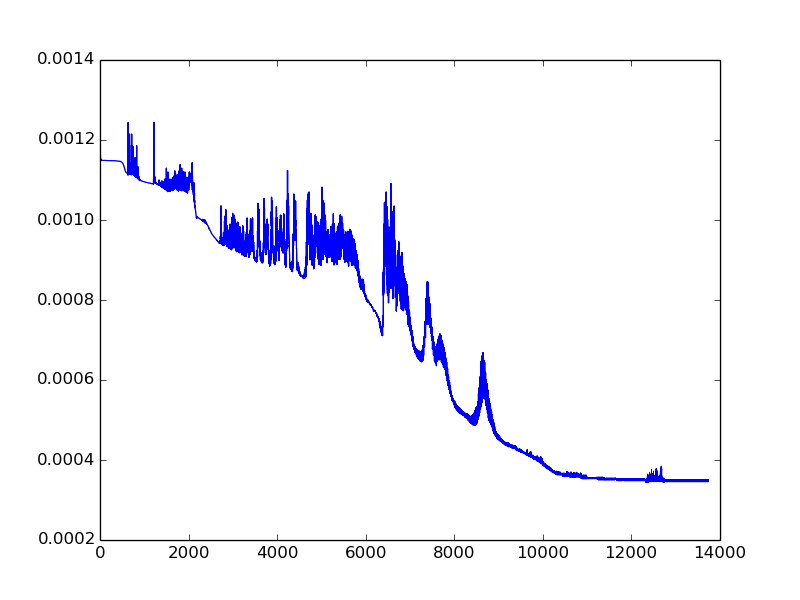}}
   \hfill
   \subfloat[Training Curve with Test-set 2\label{fig:training_curve_test3}]
   {\includegraphics[width=0.5\textwidth]{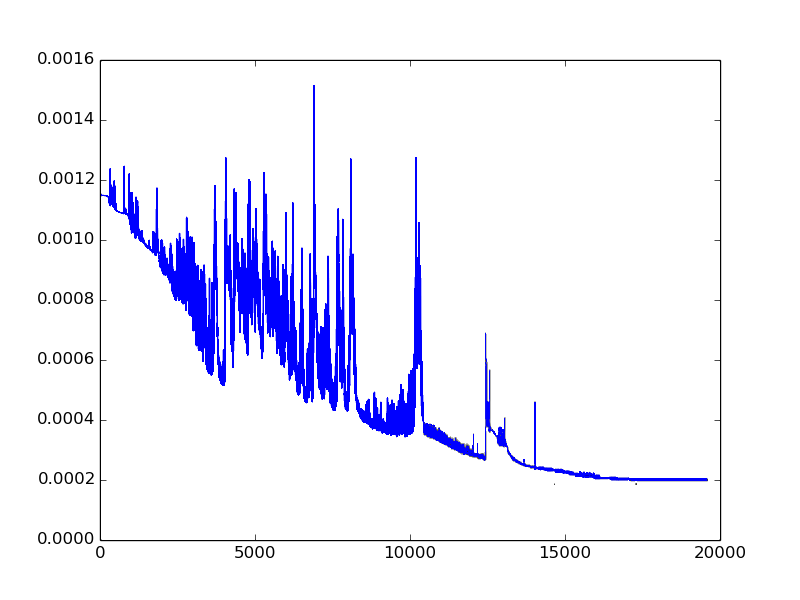}}
\hfill
\subfloat[Training Curve with Test-set 3\label{fig:training_curve_test4}]
   {\includegraphics[width=0.5\textwidth]{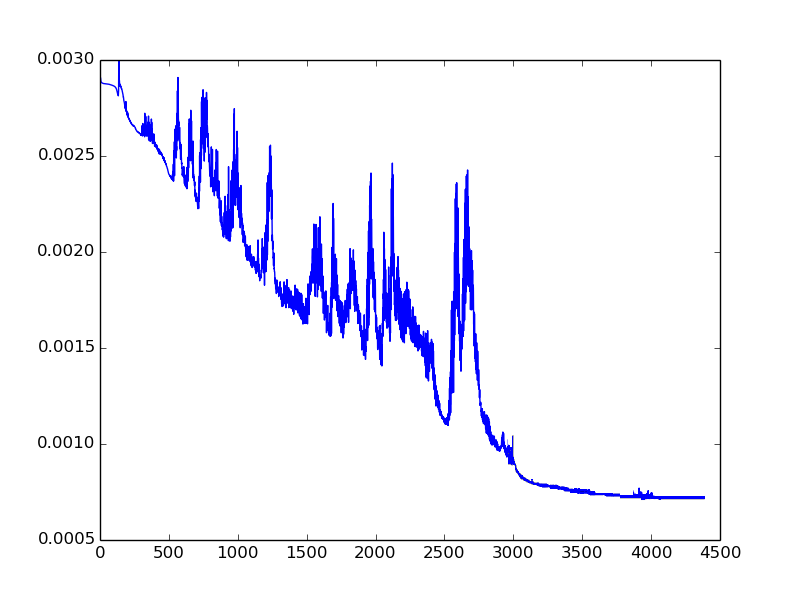}}
   \end{figure*}

\begin{table}[ht]
	\begin{center}
		\caption{RMS Error of the Generalisation Tests}
		\label{tab:rms_general}
		\begin{tabular}{|c|c|c|c|}
			\hline
			Test & 1 & 2 & 3\\
			\hline
			RMS Error (All) & 0.0052 & 0.0069 & 0.0169 \\
			\hline
			RMS Error (Step 1-20)  & 0.0064 & 0.0082 & 0.0240 \\
			\hline
			RMS Error (Step 21-40) & 0.0042 & 0.0075 & 0.0194 \\
			\hline
			RMS Error (Step 41-60) & 0.0033 & 0.0069 & 0.0150 \\
			\hline
			RMS Error (Step 61-80)  & 0.0031 & 0.0062 & 0.0121 \\
			\hline
			RMS Error (Step 81-100)  &  0.0024 & 0.0052  & 0.0101\\
			\hline
			
		\end{tabular}
		\label{tab:generalisation-error}
	\end{center}
\end{table}

\begin{figure*}[th]
  
    \caption{Trajectory Generation} The generated trajectories (dotted) with $41$ dimensions were plotted and compared with the original trajectories. Three test-sets were selected to validate the training performances with different training sets.
    \label{fig:traj-generation}
    \centering
    \subfloat[Generated Trajectory  from MTRNN 1, Test-set 61 (v.-n.: 0.1-0.1) \label{fig:traj-generation1}]
   {\includegraphics[width=0.45\textwidth]{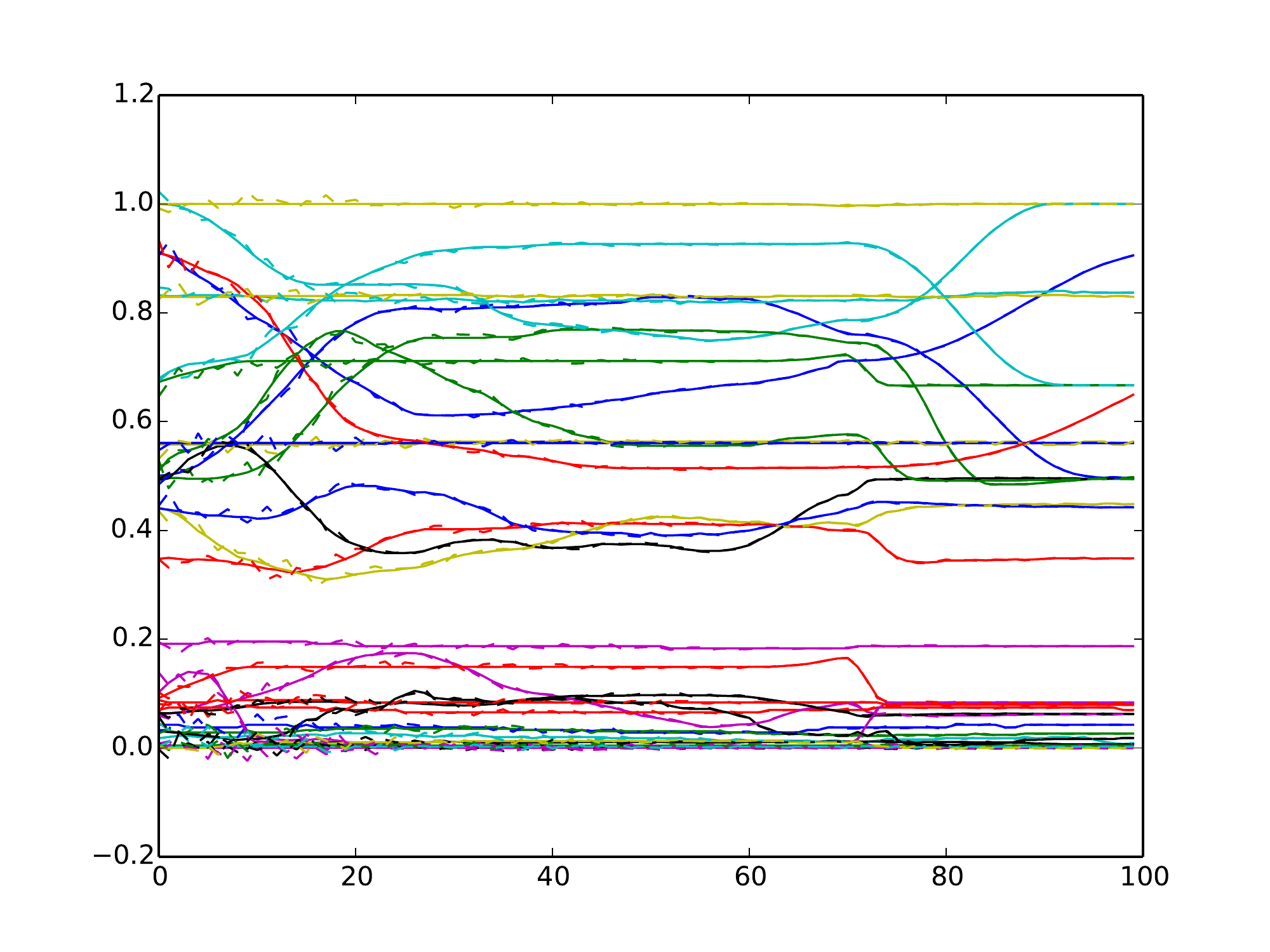}}
   \hfill
   \subfloat[Generated Trajectory  from MTRNN 2, Test-set 231 (v.-n.: 0.4-0.2) \label{fig:traj-generation2}]
   {\includegraphics[width=0.45\textwidth]{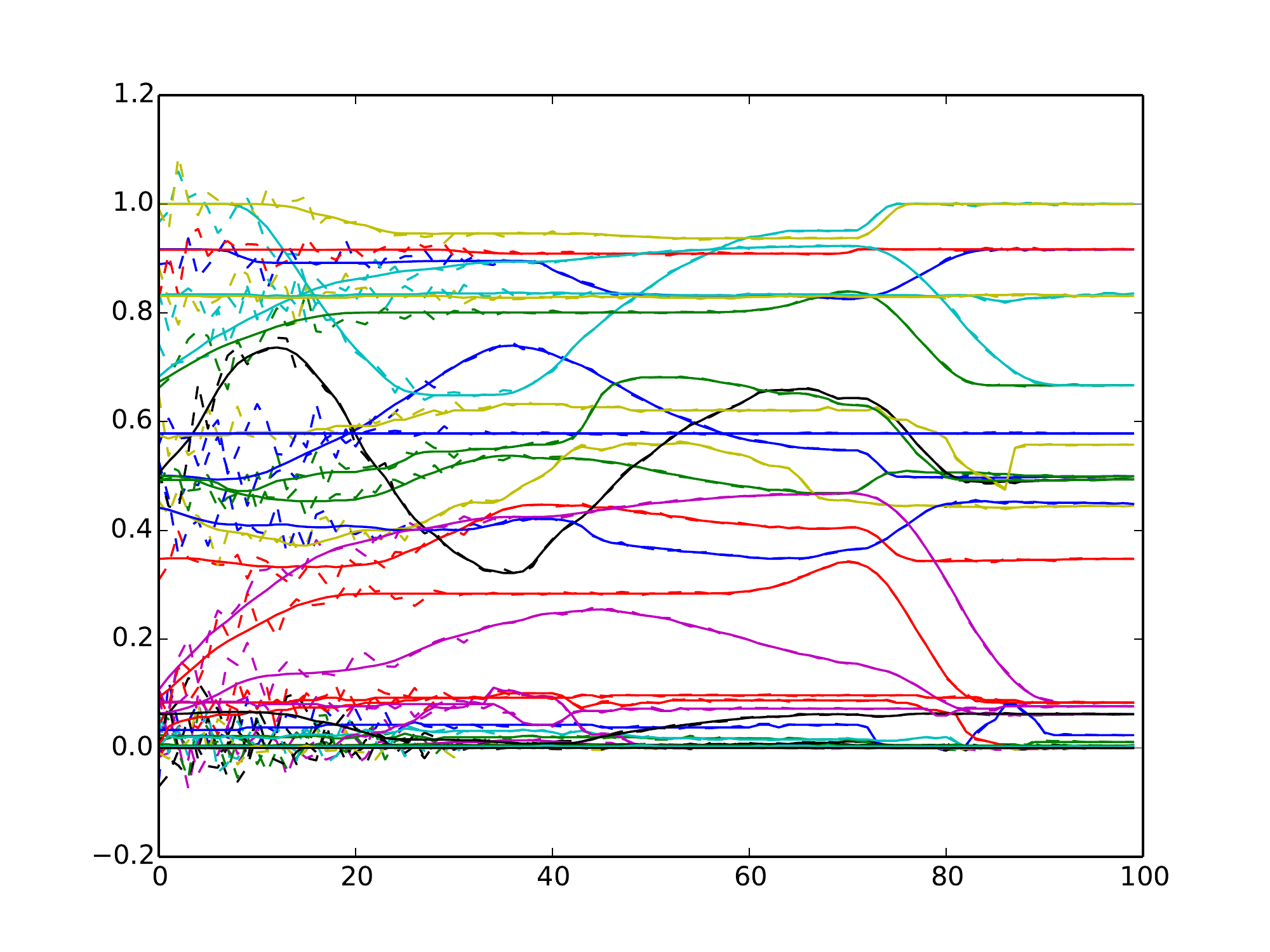}}
\hfill
\subfloat[Generated Trajectory  from MTRNN 3, Test-set 484 (v.-n.: 0.8-0.8) \label{fig:traj-generation3}]
   {\includegraphics[width=0.45\textwidth]{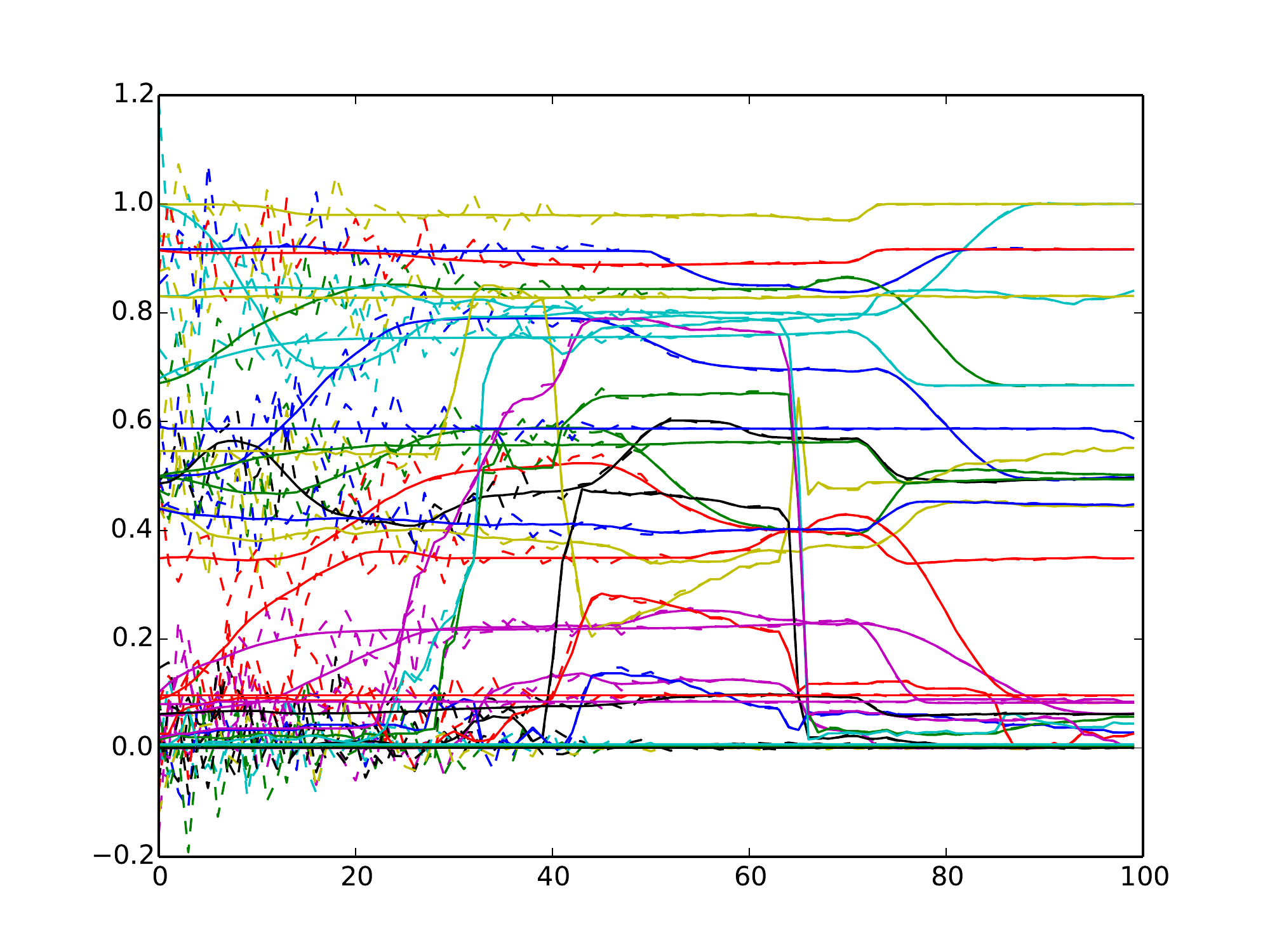}}
   \end{figure*}

Using the parameter set of $(50,5,70,100)$,  the training curves (Figs.~\ref{fig:three-training-curves}) show that the training  converged. 
To further demonstrate the robustness of the generalisation ability given the untrained sensorimotor sequences, 
the validation sets, which were not included in the training, were fed into the network. 
In this way we aimed to test how the network responds to noun-verb combinations not used during training. 
Using the three MTRNNs we trained from three data-sets,  we performed three generalisation experiments  using the missing verb-noun combinations.
In the experiments, only the first time step data in the sequence was provided (i.e. $r=0$ in Eq.~\ref{eq:feedback-rate}), which includes the initial position of the torso, head and eye motors, as well as the vocal command. Then the network prediction was used as the input of the next time-step and formed a closed-loop to complete $100$-step of the time sequence generation.  
The   errors of the whole three training-sets, as well as those in different steps are shown in Tab.~\ref{tab:rms_general}. 
A more straightforward visualisation of the network performance can be found in Figs.~\ref{fig:traj-generation}, 
which displays three examples of generated time sequences for motor actions  from three MTRNNs. 
As we calculated in Tab.~\ref{tab:rms_general}, the training error became larger when the number of training samples was smaller. In particular, a larger error could be found at the beginning of each time sequence, but the network became stable and generated a stable motor trajectory with less error as time elapsed.   
There were some errors displayed in the trajectories generation, so sometimes the generated robot behaviours based on the trajectories are biased with the original ones. However, in most of cases, the generated robot behaviours correctly followed the semantic commands~\footnote{https://youtu.be/FOgKbJ-iEhM}.

\section{Generalisation Analyses}

In this section, we focus on the problem of how the verb-noun generalisation ability of the MTRNN network is achieved. 
As shown in the previous section,  the network was able to ``understand'' the un-trained verb-noun combinations in the sense that the generated time sequences for motor actions were close to the originals, so the iCub robot can perform the action named in the vocal commands.
For an experiment with a similar aim,~\cite{sugita2005learning} reported combining two hierarchical recurrent neural networks  which can also accomplish verb-noun generalisation for understanding combinatorial semantics in a situated environment. 
The model they used, called  recurrent neural networks with parametric biases units (RNNPB), had  similar non-linear dynamics as the MTRNN: the non-linear dynamics are determined by a small number of neural units which act as bifurcation for the whole system.
 Particularly, in our case,  the learning sequences contain a much larger dimension ($35$) of the motor joint angles for the iCub movements,
   compared with motor sequences that trained in \cite{sugita2005learning}.   
   These complex sequences result in the  bifurcation which  occurs hierarchically in the MTRNN structure. 

From this point, we hypothesise that the MTRNN, or any other hierarchical RNNs, results in the separation in the network dynamics about different modalities in a self-organised way along the lexicon categories of vocal commands after training. The type of separation depends on the different organisation of the training data structures,
{This separation, with the constraints of sensorimotor sequences, occurs on different levels of the hierarchical architecture using different strategies. The way the network presents such separation in the hierarchical dynamics is self-organised, and largely depends on the data structure of the training data in the spatial and  temporal domains.  }
Particularly in our experiment setting, after enough training, the synaptic weights between a \textit{basic} motor behaviour are enforced with a particular dimension about the verb input, as it dominates a large portion of the spatio-temporal space in the sensorimotor sequences. 
The \textit{basic} motor behaviour here means that such kind of motor actions belong to general definitions such as ``slide'', ``touch'' and etc, without a specific goal for directing action. 
This is similar to the mechanism that the hearing of a verb causes in our brain: a specific area in the pre-motor cortex, corresponding to certain motor action fires when a particular verb is heard or said.  
On the contrary,  the noun also affects part of the sensorimotor outputs in terms of its role to offset the basic motor behaviours into a specific goal-directed action. 

In the following experiments, we will examine this hypothesis by means of manipulating  data and visualising the training results.

\subsection{Generalisation with Partial Inputs}

In this subsection, we concentrate on the comparisons of the results after the removal of different modalities. 
These comparisons included two parts: i) Error of generalisation after removals; ii) Visualisation of weights after removals. 

For the first part of the analysis, in order to obtain a more conclusive statement, we used two sets of data $9 \times 9$ and $3 \times 3$ of verb-noun combinations. The $3 \times 3$ data-set (Tab.~\ref{tab:remove_list_33}) contains all the combinations of three actions and three objects, which were placed into $6$ different locations.  Tab.~\ref{tab:dir-3act} was used for the vocal command discretisation. For the second part of the experiment, the visualisation of weights was only done with the $3 \times 3$ data-sets, since its features are easier to observe and its basic principle can be easily extended to the $9 \times 9$ data-set.

\begin{table}[ht]
  \caption{Dictionary of verbs and nouns for the $3 \times 3$ data sets}
  \label{tab:dir-3act}
\begin{center}
   \begin{tabular}{|c|c|c|c|}
     \hline
      Actions & Slide Left & Slide Right & Touch \\
       \hline
       Objects & Tractor & Hammer & Spikey \\
        \hline
       Values & 0.1 & 0.2 & 0.3\\
        \hline
    \end{tabular}
  \end{center}
\end{table}

\noindent For both parts of the experiment, in order to observe how different lexical categories and visual input affected  the   training results,  especially  within the output of the sequences of the motor behaviours, different parts of the input data were removed:

\begin{enumerate}
  \item No modification (base-line)
  \item Remove the noun input (i.e. the first input unit was reset to zero.)
  \item Remove the verb input (i.e. the second input unit was reset to zero.)
  \item Remove the location of the visual object (i.e.  from the third to eighth units were reset to zero.)
\end{enumerate}

\begin{table*}[h]
\begin{center}
	
		\caption{Removal of data in the $3 \times 3$ data-set. The number $i$ in the cell indicates that such a combination was removed in the $i$-th training set for the generalisation experiments. }
			\label{tab:remove_list_33}
	\begin{tabular}{|l|c|c|c|}
		\hline
		& 0&1 & 2  \\ 
		\hline
		0& {\color{red}1}  & & {\color{blue}2}   \\
		\hline
		1&   & {\color{red}1}/{\color{blue}2} &   \\
		\hline
		2& {\color{blue}2} &  & {\color{red}1}  \\
		\hline
		
	\end{tabular}

  \end{center}
\end{table*}

During the generalisation tests, the full un-trained data was placed into the network. The training and generalisation error of the motor output was compared in the Tab.~\ref{tab:remove_input_33} and Tab.~\ref{tab:remove_input_99}. 
From these two tables we can see that the removal of the verb resulted in a larger generalisation error than the other two tests, while the removal of the object location resulted in the lowest generalisation error.

\begin{table*}[h]
\begin{center}

\caption{Removal Part of Input (3 verbs and 3 nouns)}  
\label{tab:remove_input_33}
  \begin{tabular}[0.75\textwidth]{|c||c|c|c|c|c|c|}
\hline
Error & \parbox[t]{1.5cm}{w/o \\verb training}  & \parbox[t]{1.5cm}{w/o \\verb generalisation} & \parbox[t]{1.5cm}{w/o \\visual location training} & \parbox[t]{1.5cm}{w/o \\visual location training}& \parbox[t]{1.5cm}{w/o \\noun training} & \parbox[t]{1.5cm}{w/o \\noun generalisation}  \\
\hline
Test 1  &	0.0003	& 0.1041 &	0.0003	& 0.0594	& 0.0003	& 0.0868  \\

Test 2 	& 0.0003	& 0.1129	& 0.0003	& 0.0612	& 0.0003	&   0.0933\\

\hline

\end{tabular}

\end{center}
\end{table*}

\begin{table*}[h]
\begin{center}
  \caption{Removal Part of Input (9 verbs and 9 nouns)}  
\label{tab:remove_input_99}
\begin{tabular}[0.75\textwidth]{|c||c|c|c|c|c|c|}
\hline
Error & \parbox[t]{1.5cm}{w/o \\verb training}  & \parbox[t]{1.5cm}{w/o \\verb generalisation} & \parbox[t]{1.5cm}{w/o \\noun training} & \parbox[t]{1.5cm}{w/o \\noun generalisation} & \parbox[t]{1.5cm}{w/o \\visual location training} & \parbox[t]{1.5cm}{w/o \\visual location training} \\
\hline
Test 1 &	0.0003	& 0.5311 &	0.0003	& 0.5223	& 0.0003	& 0.0921 \\

Test 2	& 0.0005	& 0.6623	& 0.0005	& 0.7473	& 0.0005	& 0.1379 \\

Test 3	& 0.0006	& 0.8574	& 0.0006	& 0.7494 & 0.0006	& 0.1771	\\

\hline

\end{tabular}

\end{center}
\end{table*}

	\begin{figure*}[ht]
  
    \caption{Weight Visualization by Input Removal}
    \label{fig:remove-weight}
    \centering
    \subfloat[Weight Matrix of Normal Training (base-line)\label{fig:weight_base_line}]
   {\includegraphics[width=0.5\textwidth]{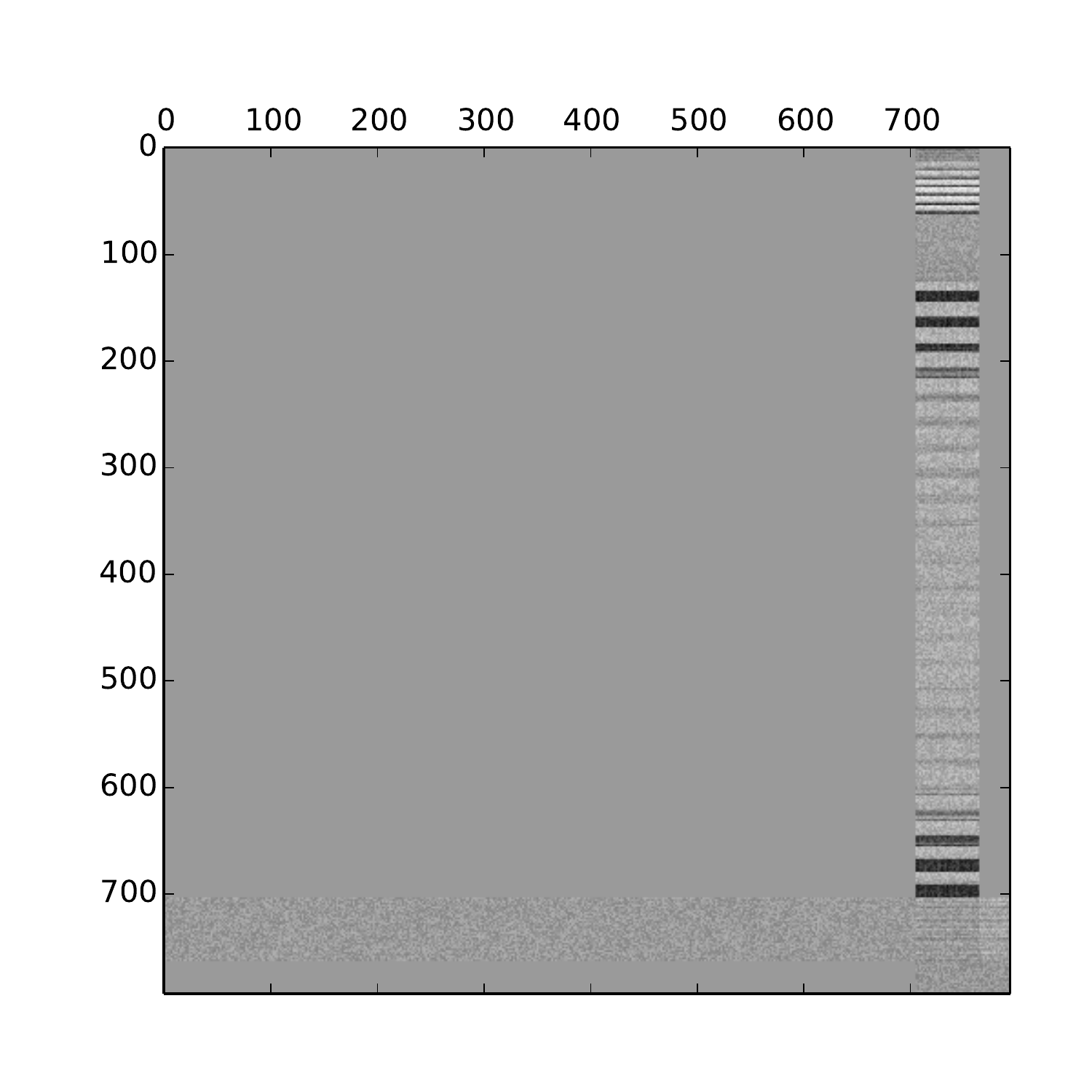}}
   \hfill
   \subfloat[Weight Matrix without Verb Input\label{fig:weight_no_verb}]
   {\includegraphics[width=0.5\textwidth]{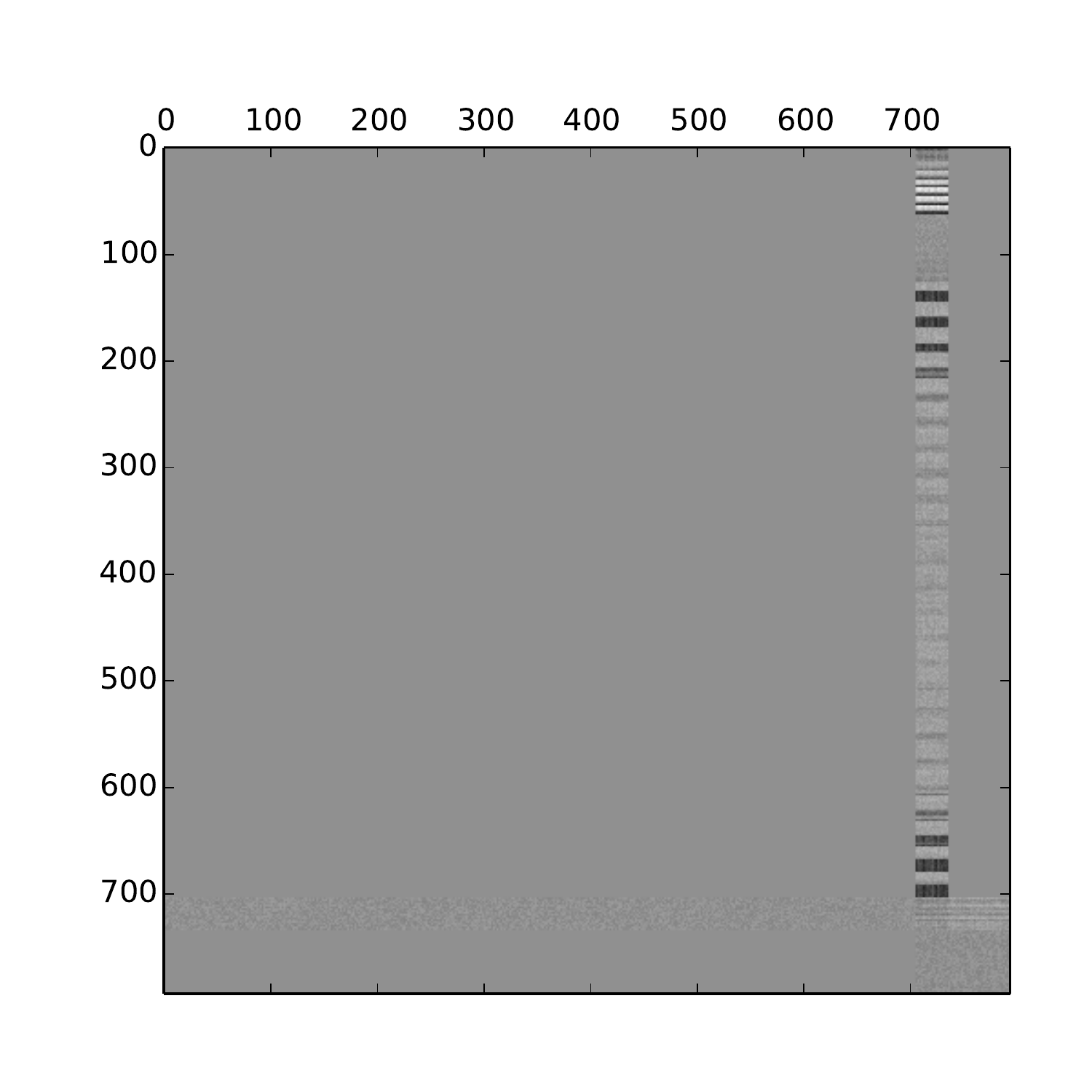}}
\hfill
\subfloat[Weight Matrix without Noun Input\label{fig:weight_no_noun}]
   {\includegraphics[width=0.5\textwidth]{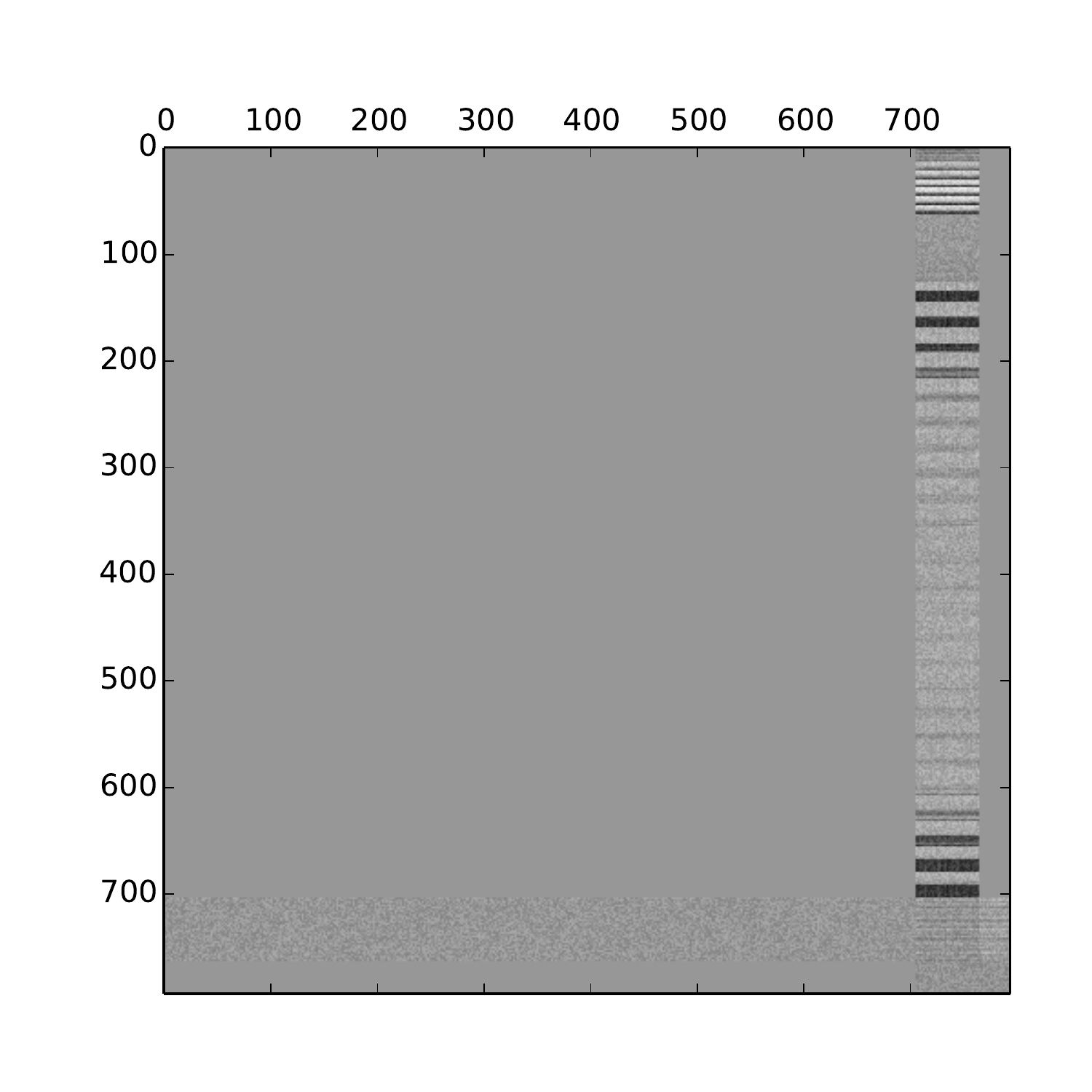}}
   \subfloat[Weight Matrix without Visual Input\label{fig:weight_no_location}]
   {\includegraphics[width=0.5\textwidth]{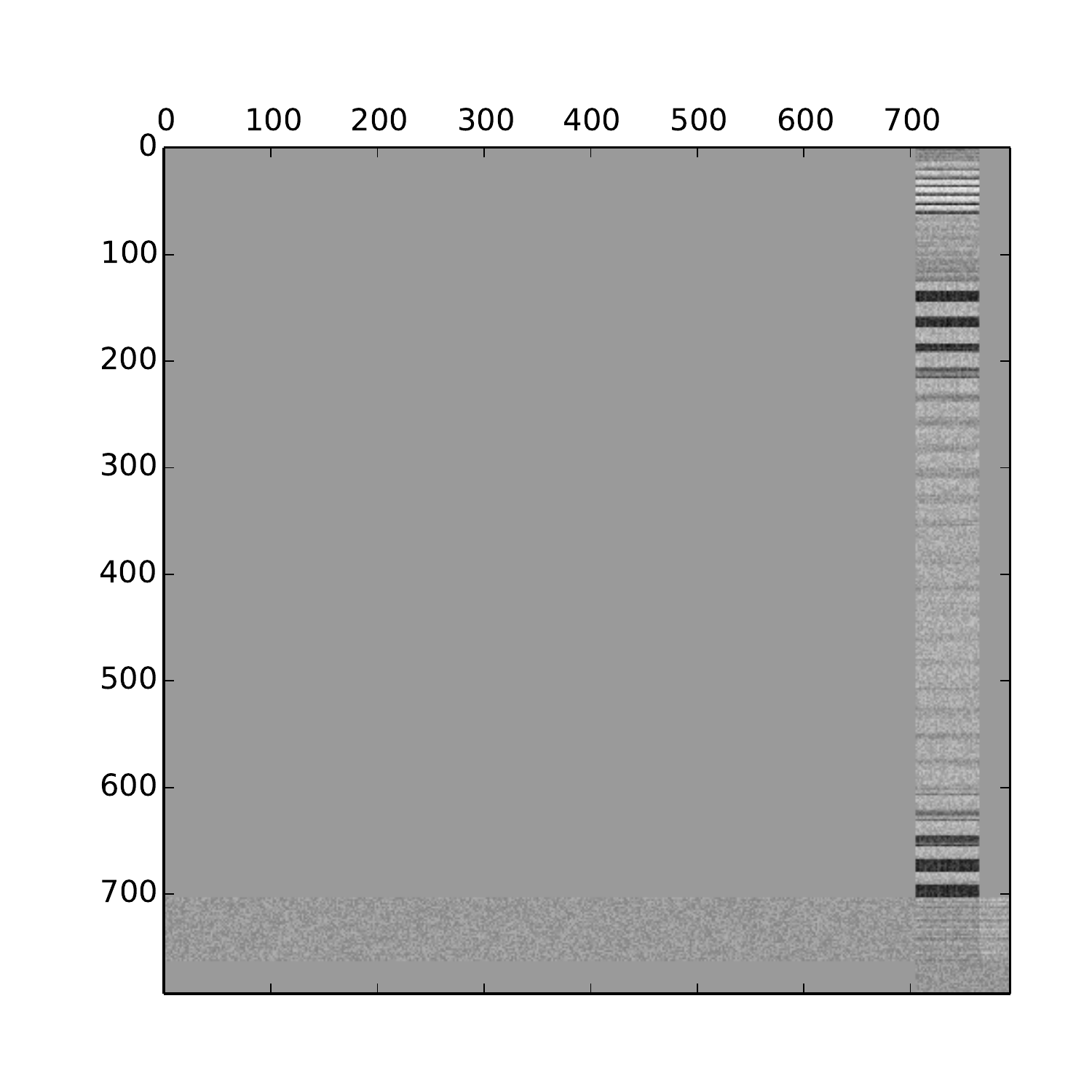}}
   \end{figure*}
	
For the second part of the experiment, the main aim was to understand the effect of a particular input modality (e.g. lexical structure and visual input) in the whole network training, by  observing the visualization of the weights.
We conducted an experiment  with a smaller data-set ($3 \times 3$) than  the previous experiments, 
due to the fact that smaller number of weights  give a better presentation for the visualization. 
But a similar conclusion would be extended into the larger $9 \times 9$ data-set. 
Figs.~\ref{fig:remove-weight} visualised the weighting matrix, where the neurons from number $0$ to number $703$ were neurons on the $IO$ layer, from number $704$ to number $764$ were neurons on the $C_f$ layer and from number $765$ to number $794$ were neurons on the $C_s$ layer. 
The weight matrices in Fig.~\ref{fig:weight_base_line}, Fig.~\ref{fig:weight_no_noun} and Fig.~\ref{fig:weight_no_location} looked quite similar. But in Fig.~\ref{fig:weight_no_verb}, without the verb input, we could easily notice that a large amount of weights from $IO$ layer to $C_f$ remain to be untrained. 
To quantitatively evaluate this observation, Tab.~\ref{tab:sim_weights} calculated the 2-norm to obtain the Euclidean distances from the manipulated weighting matrices to the base-line matrix. 
The 2-norm  was calculated by:
\begin{equation}
  d(\mathbf{W}^m - \mathbf{W}^b) = \sqrt{\sum_{i=1}^n\sum_{j=1}^n(d^m_{ij}-d^b_{ij})^2}
\end{equation}

\noindent where $\mathbf{W}^m$ is the weighting matrix after data manipulation, $\mathbf{W}^b$ is the weighting matrix from base-line experiment, $d$ is the weight from the $i$-th neuron to $j$-th neuron.  Here $n = 795 $ which is the total number of neurons.

From the comparisons of weight matrices and the Euclidean distances, we further verified our hypothesis that the lexical structure of verbs plays a significant role in the training, since it is further grounded in the differences of motor action trajectories, which dominate a large spatio-temporal space of the sequences.

	\begin{table}[th]
	  \caption{Euclidean Distances between Partial Input Matrices and Normal Training Matrix}
	  \label{tab:sim_weights}
\begin{center}

	    \begin{tabular}{|c|c|c|c|}
	    \hline
	       & w/o Verb & w/o Noun & w/o Location \\
	       \hline
	    Distance   &  8.9100 & 0.9450 & 0.6736\\
	    \hline
	    \end{tabular}
	  \end{center}
	\end{table}

\subsection{Internal Dynamics}
In the previous analysis, we have looked at the generalisation ability of the MTRNN. 
A preliminary conclusion  suggests that the lexical structure of  the verb plays a significant role in maintaining the convergence of the temporal sensorimotor sequences.
In this section we are particularly interested in  how the  generalisation
capabilities are brought by the recurrent connected hierarchical structure. 
We believed that part of these answers can be found by  observing the detailed neural activities on each context layer given the selection of different inputs. The neural activities were therefore  examined using the $9 \times 9$ data-set, with a previously trained MTRNN with the parameter setting of  $(50,5, 70,100)$.  

\begin{figure*}
\begin{center}
 \caption{Principle Component Analysis on the $C_f$ neurons} Comparison of   neural activation (after PCA) on the $C_f$ layer shows that the sequences with different nouns are clustered closer than those with different verbs. Particularly we can compare combinations of $[0, 0.0-0.3]$ and $[0.1, 0.0-0.2]$. 
\includegraphics[width=0.98\linewidth]{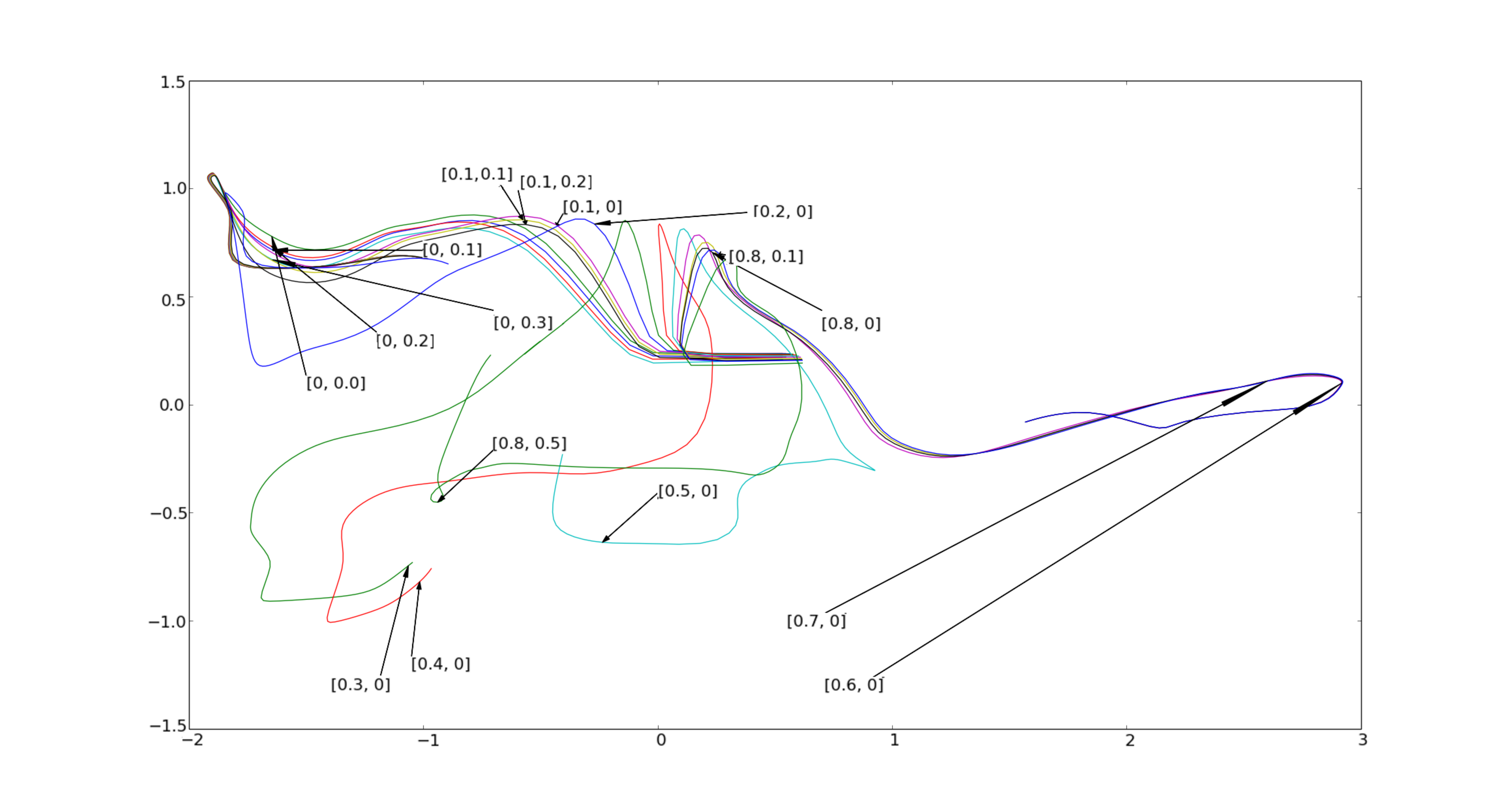}
	\label{fig:pca-c-1}
	\end{center}
	\end{figure*}

\begin{figure*}
\begin{center}
 \caption{Principle Component Analysis on the $C_f$ neurons (with different nouns): Comparison of PCA processed neural activation shows that the sequences with  different nouns differ at the beginning and at the middle of the trajectories.}  
\includegraphics[width=0.98\linewidth]{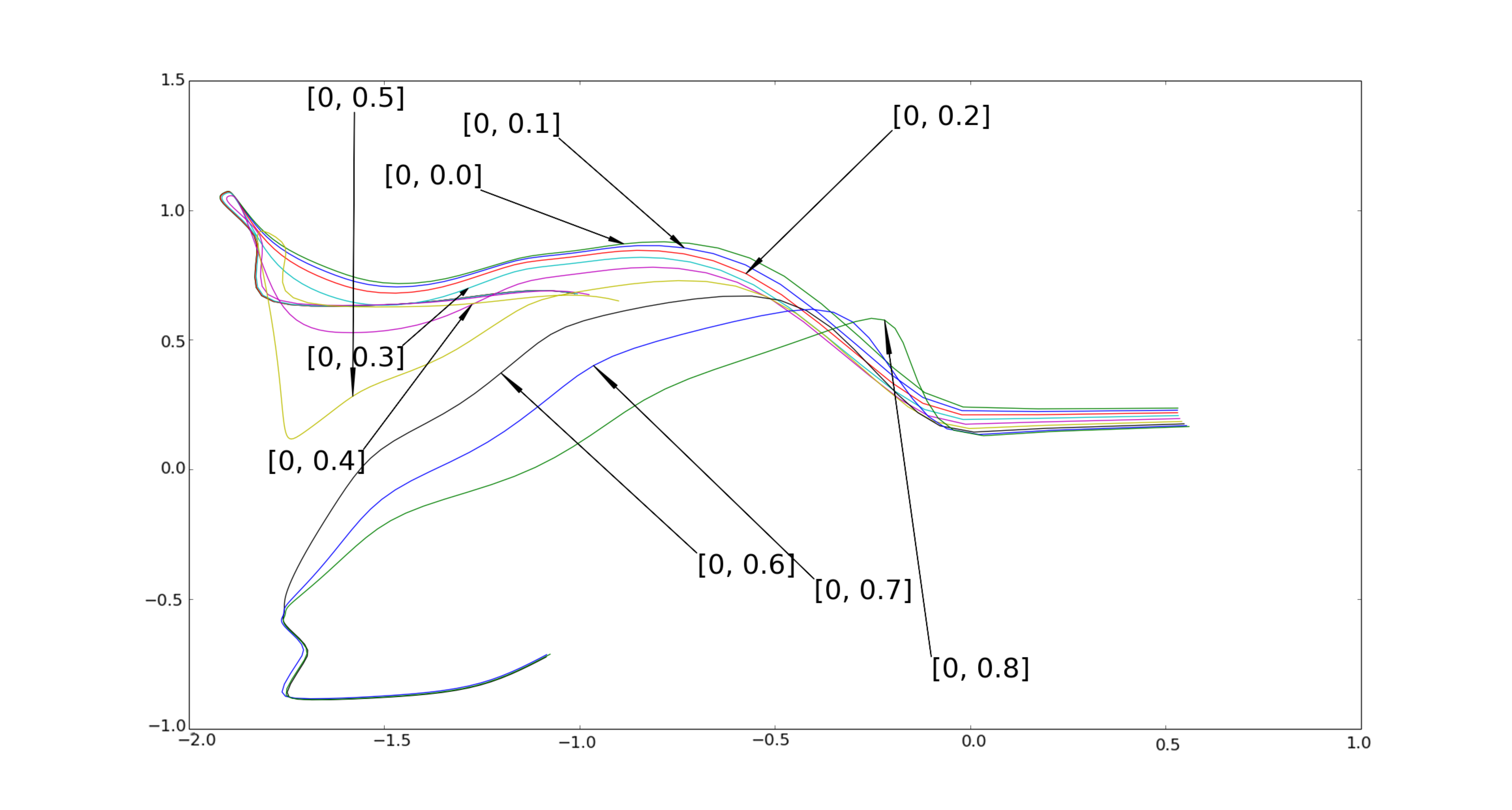}
	\label{fig:pca-c-2}
	\end{center}
	\end{figure*}

\begin{figure*}
\begin{center}
 \caption{Principle Component Analysis on the $C_f$ neurons (with different object locations): Comparison of PCA processed neural activation shows the sequences with different visual inputs result in very little divergence in the trajectories, which mainly occurs in the middle of the trajectories.}  
\includegraphics[width=0.98\linewidth]{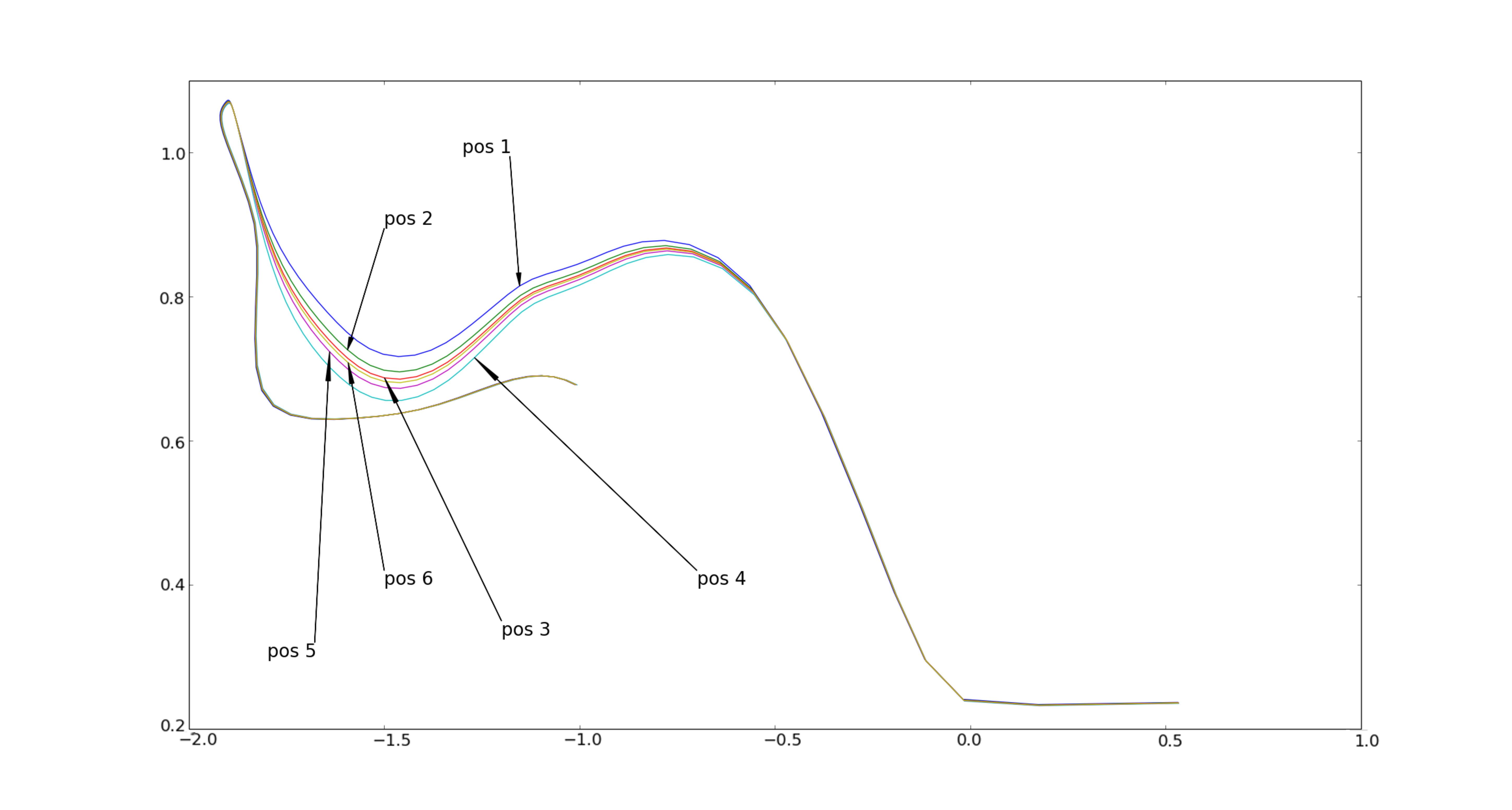}
	\label{fig:pca-c-3}
	\end{center}
	\end{figure*}

\begin{figure*}
\begin{center}
 \caption{Principle Component Analysis on the $C_s$ neurons} Comparison of PCA processed neural activation shows that the differences of the sequences are mainly present in the verbs on the $C_s$ layer.
\includegraphics[width=0.98\linewidth]{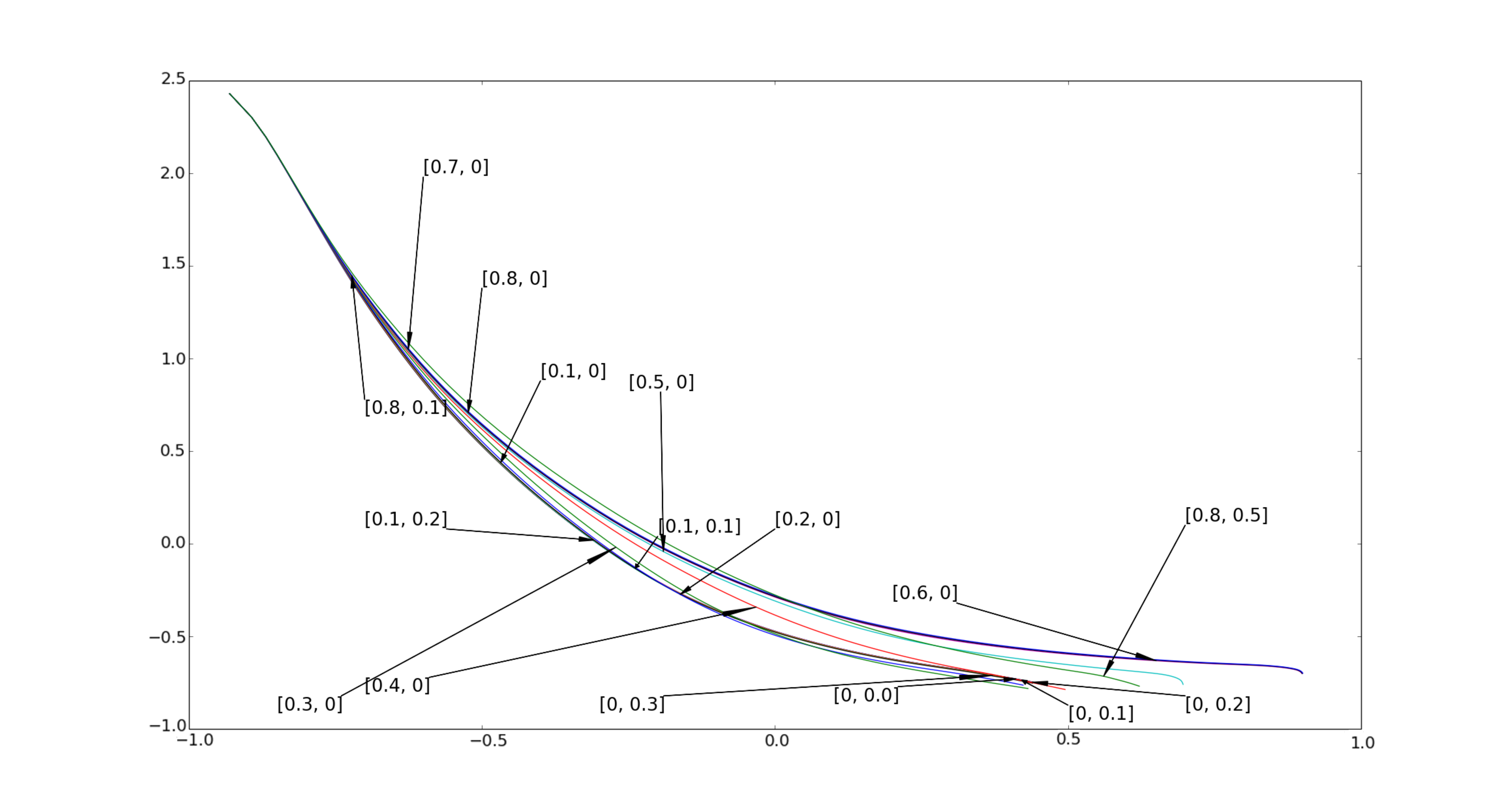}
	\label{fig:pca-c-4}
	\end{center}
	\end{figure*}

\begin{figure*}
\begin{center}
 \caption{Principle Component Analysis on the $C_s$ neurons (with different nouns)} 
\includegraphics[width=0.98\linewidth]{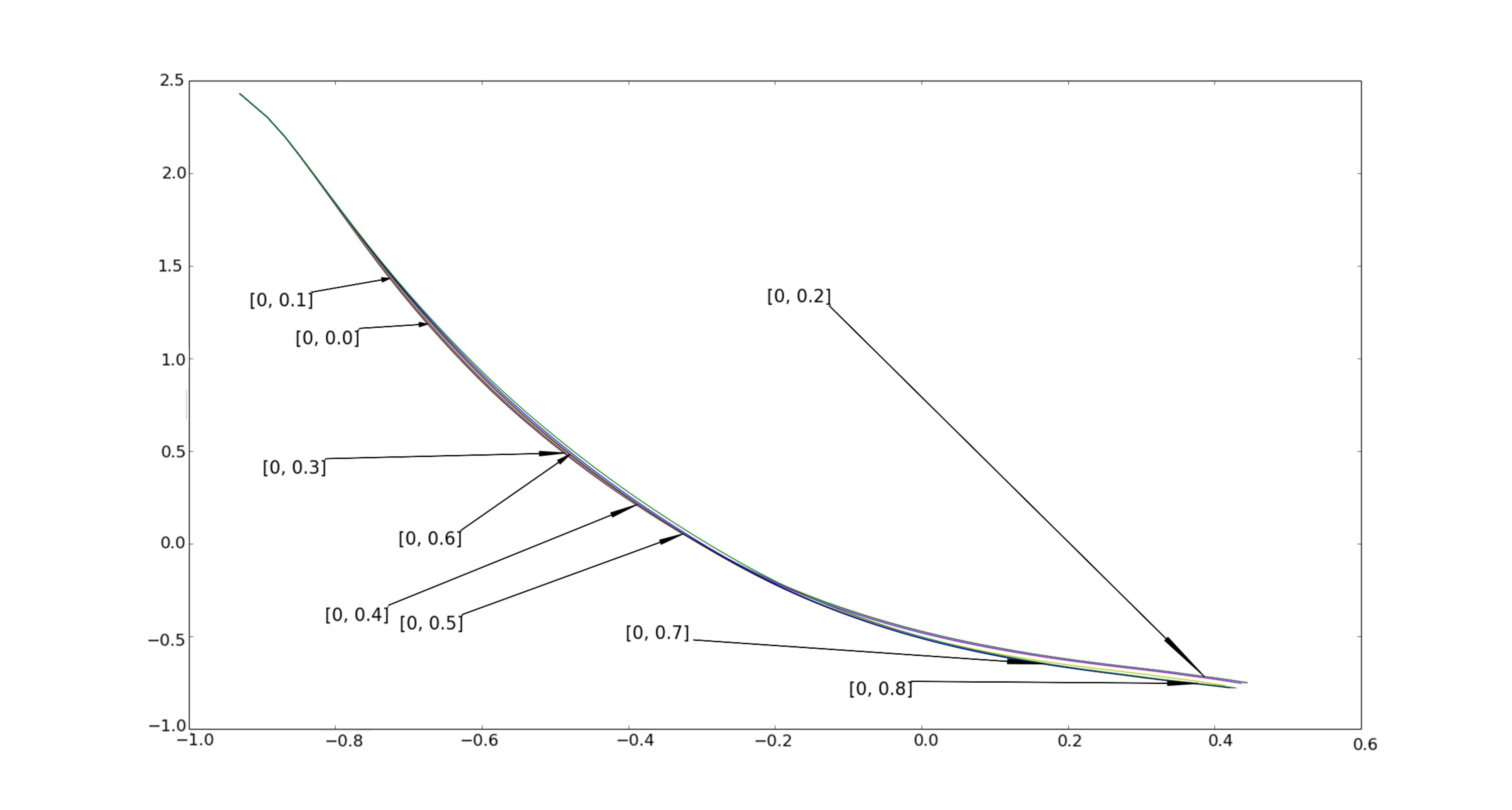}
	\label{fig:pca-c-5}
	\end{center}
	\end{figure*}

\begin{figure*}
\begin{center}
 \caption{Principle Component Analysis on the $C_s$ neurons (with different object locations)}  
\includegraphics[width=0.98\linewidth]{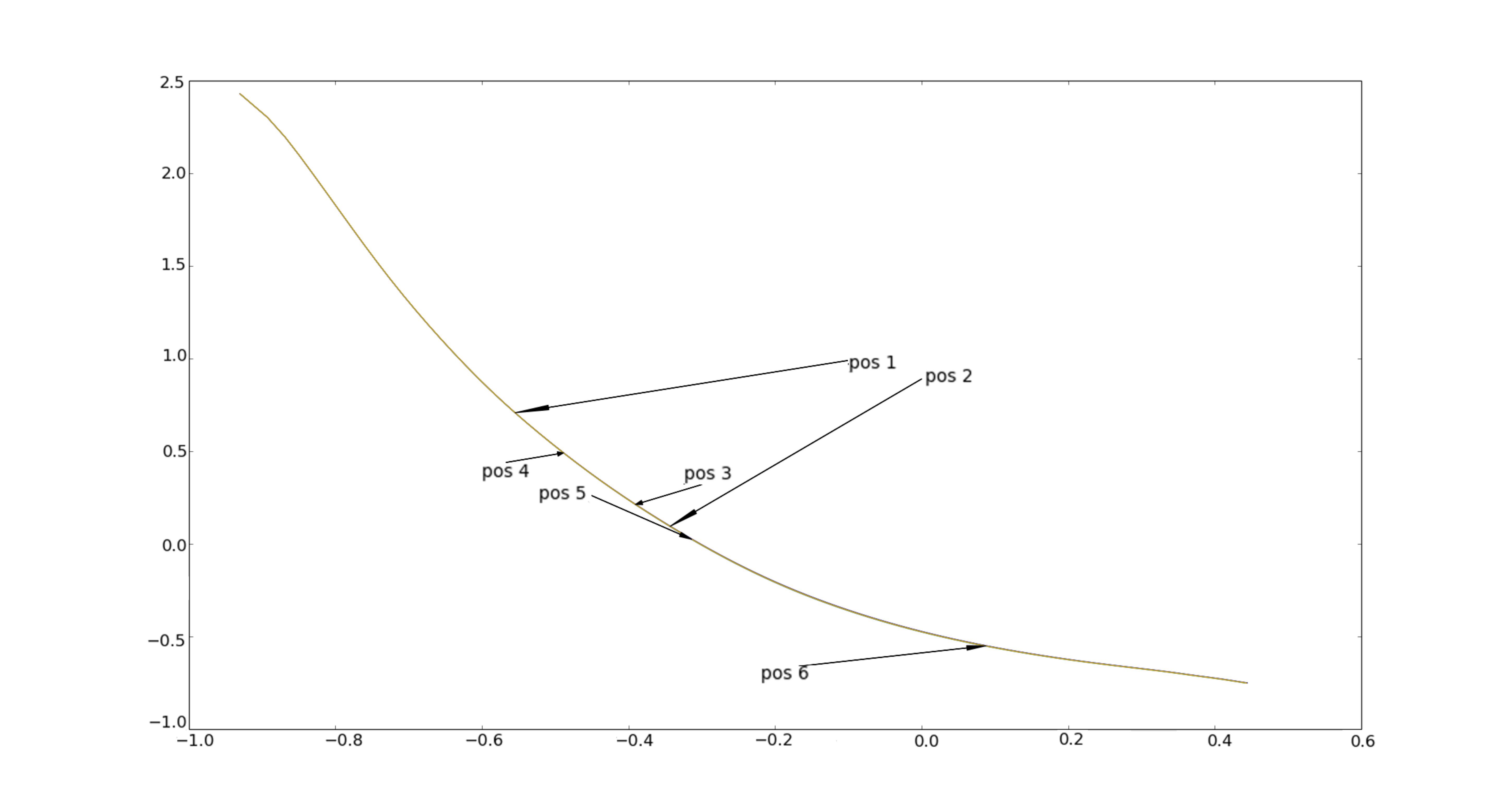}
	\label{fig:pca-c-6}
	\end{center}
	\end{figure*}

The following figures showed the PCA trajectories of the internal neural dynamics on the $C_f$ (Fig.~\ref{fig:pca-c-1}, Fig.~\ref{fig:pca-c-2} and Fig.~\ref{fig:pca-c-3}) 
and $C_s$ (Fig.~\ref{fig:pca-c-4}, Fig.~\ref{fig:pca-c-5} and Fig.~\ref{fig:pca-c-6}) layers. 
Since the complete $9 \times 9$ data-set contains $486$ sequences, whose patterns can hardly be observed in one single figure, only a few samples were presented in the following figures to clearly show the PCA trajectories. 
Fig.~\ref{fig:pca-c-1} and Fig.~\ref{fig:pca-c-4} showed the selected PCA trajectories on the $C_f$ and $C_s$ layers. These trajectories mainly concern combinations of verb inputs and a few noun inputs. 
We can see that the verbs mainly determine the patterns of the trajectories, which implies that the processing of verbs mainly affects the temporal dynamics in the MTRNN.  

The following figures mainly show how the differences in lexical structures and visual information result in the differences in the PCA trajectories. 
Fig.~\ref{fig:pca-c-2} and Fig.~\ref{fig:pca-c-5} show the PCA trajectories of the internal dynamics on $C_f$ and $C_s$ layers, with   different noun inputs; Fig.~\ref{fig:pca-c-3} and Fig.~\ref{fig:pca-c-6} showed the PCA trajectories with  different object location inputs. 
We could observe that the differences of nouns on the $C_f$ (Fig.~\ref{fig:pca-c-2}) cause divergences at the beginning of the trajectories, but not at the end. 
From Fig.~\ref{fig:pca-c-3} comparisons show the differences of visual inputs produce even smaller divergences in the trajectories, and that the divergences mainly occurred at the middle of the trajectories. 
Comparatively, from the activities on the $C_s$ layer (Fig.~\ref{fig:pca-c-5} and Fig.~\ref{fig:pca-c-6}), the divergences of the trajectories from nouns and visual inputs were even smaller: the $C_s$ layer mainly encoded the information from the verbs.

To summarise the MTRNN analysis, the model self-organises similar patterns on various levels for every sensorimotor sequence, reflecting the hierarchical  structure for the vocal commands. 
Particularly, we can see that the difference between verb inputs results in larger divergence of the trajectories than noun and object-location differences. 
Due to the data structure of our input vectors, the $IO$ layer represents a collection of each word.
With a slower adaptation rate than the $IO$ layer, the $C_f$ represents the  grounded meaning of each verb, noun and visual information.
This grounding process is learnt by all temporal sensorimotor sequences. 
Similarly, using   slower changing  neurons, the $C_s$ layer represents the general motor behaviour (i.e. the verb) of the whole sensorimotor sequence.

Therefore, the $C_f$ activation mainly represents the lexical structures (verbs and nouns). The visual location has a limited effect on the $C_f$ activation, probably because the information of noun already has overlap with the object information about visual location.
As the main factor of the $C_f$ layer, the same verbs are represented as a similar pattern on the fast context layer in all Fig. ~\ref{fig:pca-c-1}, Fig.~\ref{fig:pca-c-2} and Fig.~\ref{fig:pca-c-3}.  
 The difference from nouns can be observed at the beginning of the trajectories.  
It may correspond  to  the difference of robot behaviours at the beginning of the time sequences, caused by the neck and eye tracking before the actual hand movement starts. 
 Comparing with the $C_f$ layer, the $C_s$ activation changes even slower. It generally represents the motor behaviours; only the verbs are represented in different patterns.

\section{Discussion}

\subsection{Functional Hierarchy of RNN and its Bifurcation}
\begin{figure*}[ht]
\begin{center}
 
  \caption{A Simple Recurrent Network with Parametric Bias Units}
  \includegraphics[width=0.5\textwidth]{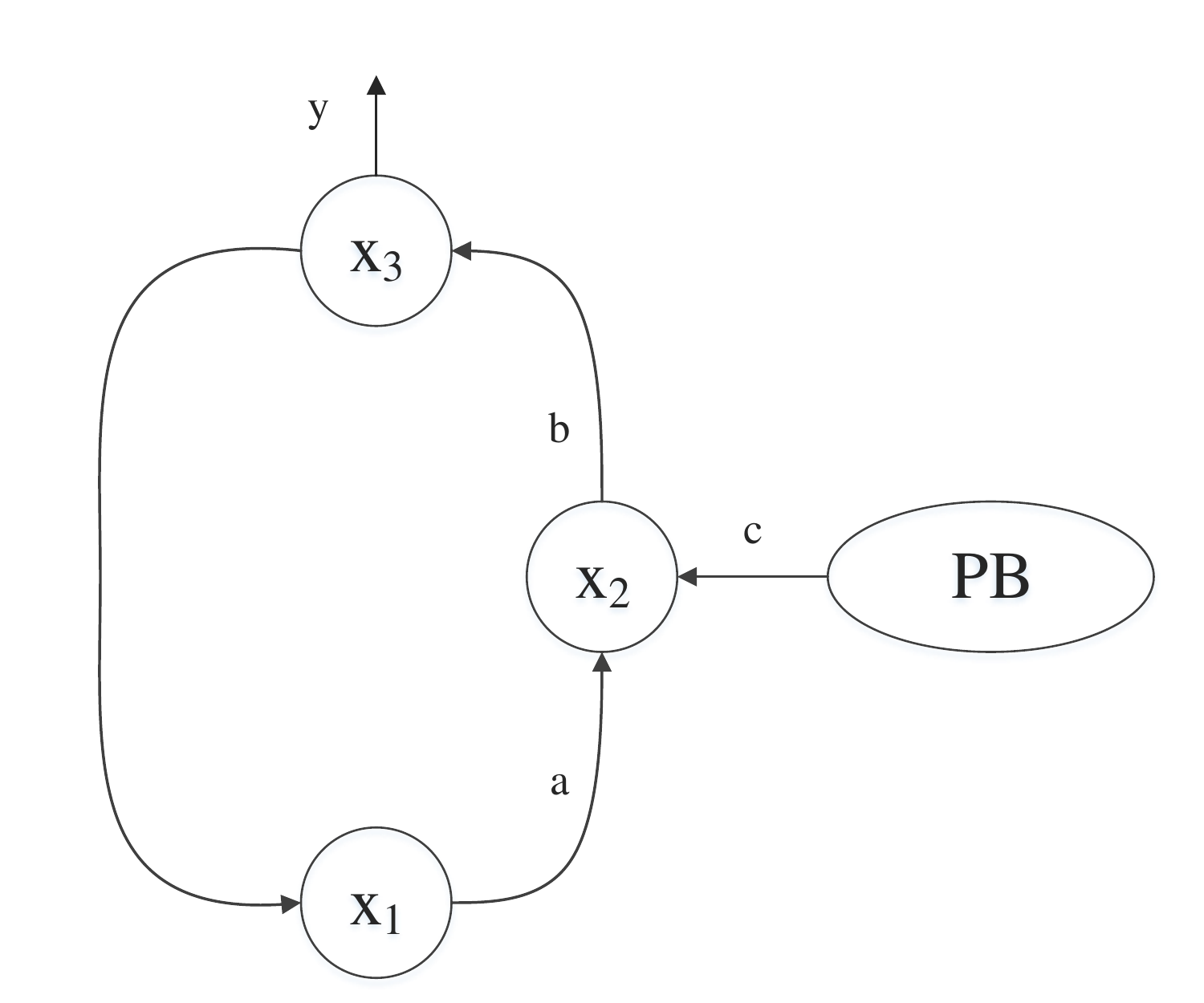}
  \label{fig:simple-pb}
   
\end{center}
\end{figure*}

It has been reported that quite a few RNN models based on functional hierarchy, such as RNNPB, MTRNN and conceptors~\cite{jaeger2014controlling}, allow the bifurcation to occur in the RNN dynamics. 
We will give a brief discussion of how this bifurcation happens.
Assuming we have a simple hierarchical RNN with an additional unit (which can be regarded as a simplified version of RNNPB) as depicted in Fig.~\ref{fig:simple-pb}. 
The system can be described as Eq.~\ref{eq:rnnpb_differential}.

\begin{eqnarray}
     \left\{
\begin{aligned}
 \dot x_{1}(t) &=& -x_1(t) + f(x_3(t)) \\
 \dot x_{2}(t) &=& -x_2(t) + a \cdot f(x_1(t)) + c \cdot PB \\
 \dot x_{3}(t) &=& -x_3(t) + b \cdot f(x_2(t)) \\
 y(t) &=& f(x_3(t))
\end{aligned}
\right.
\label{eq:rnnpb_differential}
\end{eqnarray}

There are three fixed points  in this network.  After the network has been trained, i.e. the weights $a$, $b$ and $c$ are fixed, 
 the coordinates of fixed points  only depends upon the value of PB. 
Furthermore, the coordinates of the fixed points $[x_1, x_2, x_3]$ are first-order functions of the value of $PB$ units (please see appendix for the calculation in details). 
In other words, the coordinates of the fixed points further determine the domain of different bifurcation properties. 
This is the reason that changing the parameter of $PB$ units will change the qualitative structure of the non-linear dynamics of the network. 
From the bifurcation explanation of the simplified RNNPB model, at the next step we can also extend this to  other hierarchical RNNs such as MTRNN, as they are holding a fundamentally similar theoretical foundation~\cite{tani2014self}.
  
\subsection{Generalisation Ability of MTRNN}

In our  experiments, the MTRNN was trained under a particular input data structure: 
Firstly the language commands were recorded as auditory data and transformed into a discrete  symbolic representation, and secondly the object locations and the motor behaviours were also stored as the angles of motor joints. 
This unique structure is a simplified representation of the common coding theory, which proposes that perceptual inputs and motor actions are sharing the same format of the representation within the cognitive processes. 

The neural dynamics in our MTRNN exhibited a dynamics which are different from those reported in~\cite{hinoshita2009emergence} and  \cite{heinrich2015analysing}. Whereas the noun (or object perceptual inputs) play  a  significant factor in the dynamics of context layers in these two examples, our network has minimised the effects of nouns or the object perception. 
This is partly because the input data structure, where the motor joints of the iCub robot have much larger dimensions than the  visual perception input. 
Also the spatial information for objects in our experiment setting is much easier to learn, compared with our diversified motor behaviours.
The generalisation here concerns more  the inference of symbolic meaning of a language command 
due to the composition of neural dynamics. 
During the training in a hierarchical network, such as MTRNN or RNNPB,  the neural connections strengthen between a particular type of sensorimotor sequence (motor angle changes due to differing behaviours) and  visual perception.
Particularly, in our case of $9 \times 9$ data-sets, most of our network weights stores the learning of motor actions.
 
Note that the generalisation of commands in the verb-noun combinations is not the same as we usually do in the generic recurrent neural networks (e.g.~\cite{ito2004generalization,pineda1987generalization,zhong2014toward}), which expect the network to do interpolation or extrapolation with a novel input value in either temporal or spatial space. 
While generalizing dynamical patterns by interpolation is a non-trivial task for training motor patterns in robots, 
 our main concern is the novel combinations in the context of  lexicon acquisition. 
In our case, the learning of verbs and nouns results in the emergence of different dynamics that are mostly stored in different synaptic weights, and thus their combinatorial composition is realised by the non-linearity of the recurrent connections. 
Considering the different generalisation abilities of generic RNN, RNNPB~\cite{kleesiek2013action,zhong2014toward} and MTRNN~\cite{heinrich2015analysing}, the
hierarchical RNNs appear particularly suitable for the production of flexible  motor behaviour and language expression simultaneously in the real-world social robot experiments. 

\subsection{Thought Vectors and Further Development}
A few machine learning methods have  recently  been proposed based on the encoder-decoder (ED) architecture~\cite{cho2014learning}, which  achieved great performance in machine translation~\cite{sutskever2014sequence}, image captioning~\cite{vinyals2014show}, etc. 
The ED architecture usually consists of two recurrent neural networks. One deep RNN network encodes a sequence of input vectors with arbitrary length into a fix-length vector representation in a hierarchical way,
while the other  deep RNN network decodes this representation into a  target sequence of output vector.
This specific representation between the encoder and the decoder RNNs is called ``thought vectors'' which is claimed to represent the meaning of the sequence in a high-dimensional space. 
The training of such an architecture is done by maximizing the conditional probability of the target sequence. 
If the input sequence is denoted as $(x_1, x_2, \cdots, x_T)$ and the corresponding output sequence is $(y_1, y_2, \cdots, y_{T'})$ ($T$ does not necessarily equal to $T'$), the  next symbol generation is done by maximising Eq.~\ref{eq:prob}.

\begin{strip}
\begin{equation}
	\centering
 \prod_{t=1}^{T'} P(y_t|y_{t-1},y_{t-2},\cdots,y_{1},c) = P(y_{T'},y_{T'-1},\cdots,y_{1}|x_{T},x_{T-1},\cdots,x_{1})
 \label{eq:prob}
\end{equation}
\end{strip}

\noindent Generic RNNs are not able to approximate the probability of the  sequence with arbitrary length because of its vanish gradient problem,
but other novel RNNs, such as LSTM, BRNN (Bi-directional Recurrent Neural Networks), have been successfully  employed to construct the ED architecture to ``understand'' (encode) and to ``generate'' (decode) the temporal sequences. 
Furthermore, due to the recent popularity of parallel computation by GPU, it has become possible to train and use such architectures to solve   problems such as machine translation and image captioning.

As the MTRNN can also avoid the vanish gradient problem, and larger MTRNN can be implemented via GPU, it is also possible to embed the MTRNN into the ED architecture. 
In fact, the context slow level $C_s$ already exhibits a similar feature of ``thought vectors'', using a stable neural vector to represent the basic profiles of motor actions and object instances (in our robotic experiment).
They also have similar information bi-directional flows which allow the networks to recognise and to generate the time sequences. 
Despite  their similarities, compared with LSTM, the MTRNN have other distinct features: First, from the above experiments and from other MTRNN experiments~\cite{heinrich2015analysing,hinoshita2009emergence}, it has been shown that the fast context layers and slow context layers exhibit various dynamics to explicitly represent the relationship between the verbs and nouns. The deep LSTM, on the contrary, has not been reported to have similar dynamics.
Second, differently from the static vector representation from LSTM, the context layers allow a ``slow'' change through time which is more realistic for an interaction environment, where it can be used to dynamically exhibit the meaning of sentences and sensorimotor information.

Admittedly, the training of deep RNNs, e.g. LSTMs and MTRNNs, costs a large amount of computational effort. But the recent development of GPU computing provides an opportunity to construct and test such a big scale neural network with a reasonable time and budget. 
The combination of MTRNN, the concept of ``thought vectors'' and its embodiment in robotic systems, will allow us to further explore issues such as:
\begin{enumerate}
  \item The comparison of the performances of MTRNN, LSTM and BRNN within the ED architecture and examine their performances in the robotic platforms.
  \item The robot motor action, as  a natural temporal sequence, can be further incorporated as the training of RNNs of ED architecture with connections to other modalities.
\end{enumerate}
 
\section{Conclusion}
This paper presents a neurorobotic study on noun and verb generation and generalisation, utilising with the MTRNN networks, 
with a large data-set, consisting of vocal language commands, visual object and motor action data. 
Although the generalisation abilities of hierarchical RNNs (RNNPB, MTRNN) have been reported in previous research, 
 this is the first study to demonstrate its generalisation capability using such a large data-set, which enables the robot to learn to handle real-world objects and actions.  
These experiments showed that the generalisation ability of the network are possible even with a large amount of test-sets ($9$ motor actions and $9$ objects placing placed in $6$ different locations). 
This is particularly important because  the recurrent connections between the verbs and nouns are associated with different modalities of the training-data, which is strengthened during embodiment training by the sensorimotor interaction.
Detailed analyses on the robot's neural controller showed that   the dynamics on different layers are self-organized in the MTRNN.
These self-organised dynamics further constitute a functional hierarchical representation on different layers, which associate different lexical structures with different modalities of the sensorimotor inputs. 
The MTRNN showed how the embodied information about the verbs  dominates a large portion of the network dynamics, since the proprioception information plays a significant role in the training sequences.  
As such, the hierarchical RNNs, such as MTRNN, are shown to be particularly beneficial in building a neurorobotics cognitive architecture about language learning for robotic systems, where the recurrent connections are able to self-organise and build   associations between embodied information in different modalities and the lexical structure information.

\begin{acknowledgements}
	This research has been  supported by the EU project POETICON++ under grant agreement 288382 and the UK EPSRC project BABEL. 
	We are grateful to Dr. Christopher Ford for his helpful review. 
	
\end{acknowledgements}

\bibliography{MTRNN}
\bibliographystyle{plain}

\section*{Appendix}

To calculate the coordinates of the fixed points, we should let $f'(x)=0$, which means that we need to solve the following equations

 \begin{eqnarray}
 \centering
     \left\{
\begin{aligned}
   -x_1(t) + f(x_3(t)) = 0 && \\
   -x_2(t) + a \cdot f(x_1(t)) + c \cdot PB = 0 && \\
   -x_3(t) + b \cdot f(x_2(t)) = 0 && \\
\end{aligned}
\right.
\label{eq:rnnpb_fixed_point}
\end{eqnarray}

The first solution for the first coordinate $[x_1^1, x_2^1, x_3^1]$  is:


\begin{eqnarray}
\centering
  \left\{
\begin{aligned}
 x_1^1  &=   \sqrt{\frac{36N^2+(6M-6N^2+Na)^2}{36N^2b^2+36N^2+(6M-6N^2+Na)^2}} \\
 x_2^1  &=  \frac{a}{6} + \frac{M}{N} - N  \\
 x_3^1 &=  6 \sqrt{\frac{N^2 b^2}{36N^2+(6M+N(-6N+a))^2}} \\
\end{aligned}
\right.
\label{eq:simple_PB_sol_1}
\end{eqnarray}

 Similarly,  the coordinate of the second fixed point $[x_1^2, x_2^2, x_3^2]$ is calculated by:

\begin{strip}
	\\ \\
\begin{eqnarray} \left\{
\begin{aligned} 
& \scriptstyle x_1^2  \scriptstyle =  \sqrt{\frac{ [2250000(173N-100)^2+(-30000M+224727N^2+43250Na +259650N+25000a+75000)^2]}   {[{225000b^2(173N-100)^2+225000(173N-100)^2+(-30000M+224727N^2 +43250Na+259650N+25000a+75000)^2}] }} \\
& \scriptstyle  x_2^2  \scriptstyle  = \scriptstyle  \frac{a}{6} + \frac{2M}{-1+\sqrt{3}}N-(-\frac{1}{2}+\frac{\sqrt{3}}{2}\cdot N) && \\
& \scriptstyle x_3^2 = \scriptstyle 6\sqrt{\frac{b^2\cdot (1.73N - 1.0)^2}{-12M + (1.73N - 1.0)\cdot(5.196N + a + 3.0)^2 + 36(1.73N - 1.0)^2}} &&  \\
\end{aligned}\right. 
\label{eq:simple_PB_sol_2}
\end{eqnarray}
\end{strip}

And the coordinate of the third fixed point $[x_1^3, x_2^3, x_3^3]$ is given by 

\begin{strip}
\begin{eqnarray} \left\{
\begin{aligned}
& \scriptstyle x_1^2  \scriptstyle =  \sqrt{\frac{ [2250000(173N+100)^2+(-30000M+224727N^2+43250Na +259650N+25000a+75000)^2]}   {[{225000b^2(173N+100)^2+225000(173N+100)^2+(-30000M+224727N^2 +43250Na+259650N+25000a+75000)^2}] }} \\
& \scriptstyle  x_2^2  \scriptstyle  = \scriptstyle  \frac{a}{6} + \frac{2M}{-1-\sqrt{3}}N-(-\frac{1}{2}-\frac{\sqrt{3}}{2}\cdot N) && \\
& \scriptstyle x_3^2 = \scriptstyle 6\sqrt{\frac{b^2\cdot (1.73N + 1.0)^2}{-12M + (1.73N + 1.0)\cdot(5.196N + a + 3.0)^2 + 36(1.73N + 1.0)^2}} &&  \\
\end{aligned}\right. 
\label{eq:simple_PB_sol_3}
\end{eqnarray}
\end{strip}

\noindent For the above solutions, we define the parameters $M$ and $N$ as follows:

\begin{eqnarray*}
  M &=& -\frac{a^2}{36} - \frac{b^2}{6} - \frac{1}{3} \\
N &=& \Bigg[ -\frac{a^3}{216} + \frac{ab^2}{4} + \frac{-\frac{ab^2}{2} - 1}{12}  + \frac{a}{4} + \frac{c \cdot PB}{4} \\ && + \sqrt{M^3 + \frac{(-\frac{a^3}{108} + \frac{ab^2}{2} + \frac{ -\frac{ab^2}{2} - 1}{6} + \frac{a}{2} + \frac{c \cdot PB}{2})^2}{4}} \Bigg]^{1/3}
\end{eqnarray*}

\noindent Although the equations seem to be complicated, remember that variables $a$, $b$ and $c$ (weights) are constant after training, which means that $M$ is a constant as well. Thus $PB$ value is a first-order variable in the function of $N$. 
Similarly, from observation from Eqs.\ref{eq:simple_PB_sol_1}  - \ref{eq:simple_PB_sol_3} we can see that the solutions are first-order function of variable $N$, which means that the coordinates of this non-linear system are a first-order function of $PB$.

\end{document}